\documentclass[11pt]{article}

\usepackage[final]{acl}

\usepackage{times}
\usepackage{latexsym}
\usepackage{amsmath}
\usepackage{amssymb}
\usepackage{mathtools}
\usepackage{amsthm}
\usepackage[T1]{fontenc}

\usepackage[utf8]{inputenc}

\usepackage{microtype}
\usepackage{microtype}
\usepackage{graphicx}
\usepackage{subcaption}
\usepackage{booktabs} 
\usepackage{multirow} 

\usepackage{inconsolata}

\usepackage{enumitem}
\usepackage{tcolorbox}
\usepackage{nicefrac}

\newcommand{\ac}{a(q,c)}
\newcommand{\ap}{a(q,\phi)}
\newcommand{\codeurl}{\url{https://github.com/copenlu/pk-ck-knowledge-disentanglement}}

\usepackage[capitalize,noabbrev]{cleveref}

\newtheorem{theorem}{Theorem}[section]

\title{Multi-Step Knowledge Interaction Analysis via Rank-2 Subspace Disentanglement}

 \author{
Sekh Mainul Islam,~
Pepa Atanasova,~
Isabelle Augenstein
\\
University of Copenhagen \\
\texttt{\{seis, pepa, augenstein\}@di.ku.dk}
}

\begin{document}
\maketitle
\begin{abstract}
  Natural Language Explanations (NLEs) describe how Large Language Models (LLMs) make decisions by drawing on external Context Knowledge (CK) and Parametric Knowledge (PK). Understanding the interaction between these sources is key to assessing NLE grounding, yet these dynamics remain underexplored. Prior work has largely focused on i) single-step generation and ii) modeled PK--CK interaction as a binary choice within a rank-1 subspace. This approach overlooks richer interactions and how they unfold over longer generations, such as complementary or supportive knowledge. We propose a novel rank-2 projection subspace that disentangles PK and CK contributions more accurately and use it for the first multi-step analysis of knowledge interactions across longer NLE sequences. Experiments across four QA datasets and three open-weight LLMs demonstrate that rank-1 subspaces struggle to represent diverse interactions, whereas our rank-2 formulation captures them effectively, highlighting PK alignment for supportive interactions and CK alignment for conflicting ones. Our multi-step analysis reveals that hallucinated generations exhibit strong alignment with the PK direction, though context-faithful generations maintain a more balanced alignment between PK and CK\footnote{Code and data: \codeurl{}}.
\end{abstract}

\begin{figure*}[t]
    \centering
    \includegraphics[width=\linewidth]{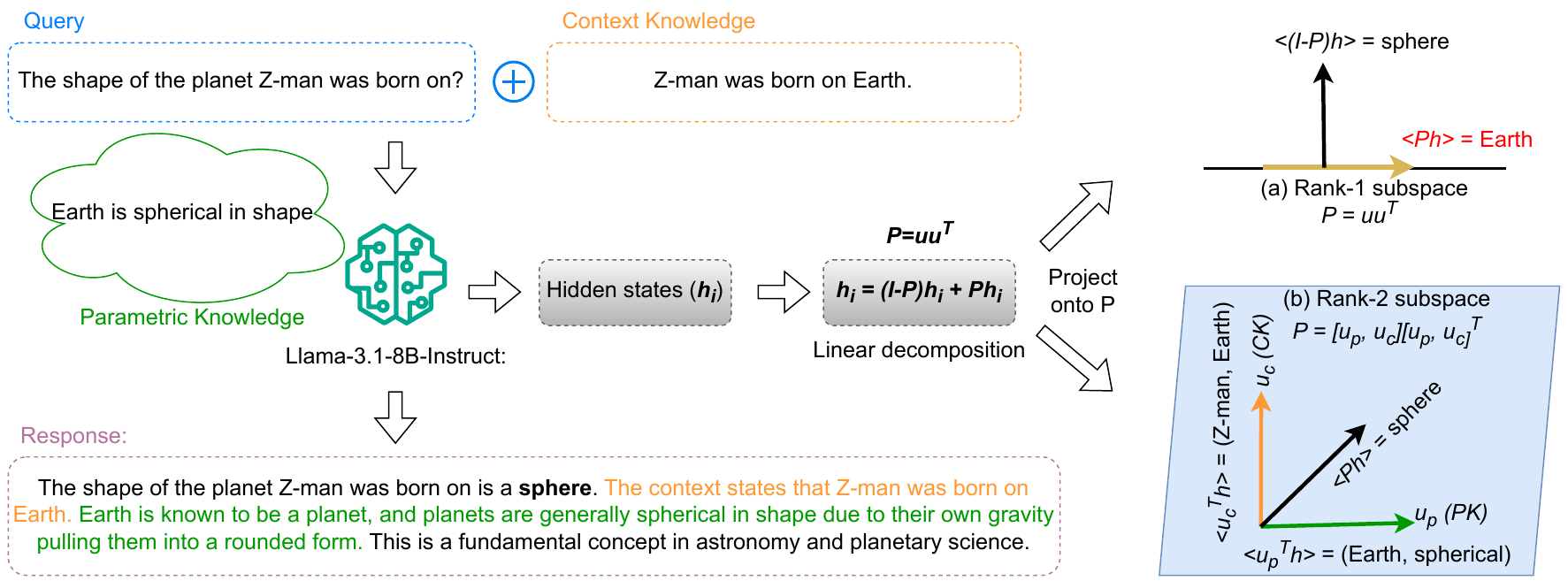}
    \caption{(Left) Llama-3.1-8B-Instruct model combines parametric (green) and contextual (orange) knowledge to generate NLEs. (Right) Projection of the answer token to a learned low-rank subspace $\mathbf{P}$ disentangles its individual knowledge contributions -- (a) rank-1 fails with the correct knowledge captured only in its orthogonal subspace, whereas (b) rank-2 succeeds in encoding the correct knowledge sources along the disentangled axes.}
    \label{fig:overall_diagram}
\end{figure*}

\section{Introduction}\label{sec:introduction}

Large Language Models (LLMs) are employed to generate Natural Language Explanations (NLEs) in a human-readable format, illustrating the underlying decision-making process for predictions in complex reasoning tasks such as Claim Verification (CV) and Question Answering (QA). These NLEs are valuable as they can reveal the utilization of external Context Knowledge (CK) and the Parametric Knowledge (PK) of a model. Consider the \cref{fig:overall_diagram} example from a QA task \citep{cheng2024understandinginterplayparametriccontextual}, where Llama-3.1-8B-Instruct \cite{grattafiori2024llama3herdmodels} generates the NLE by utilizing both CK and PK, to explain the underlying decision-making process for the final answer prediction. For some tasks, such as CV, we assume the decision-making process and, in turn, the NLE, will rely more on evidence (CK) \cite{wang-shu-2023-explainable,tan-etal-2025-improving}, and more on PK in QA tasks with misleading external context. On the other hand, Chain-of-Thought (CoT) prompting \citep{wei2022chain}, a widely adopted reasoning methodology, explicitly elicits NLEs of the intermediate reasoning steps and is hypothesized to influence how LLMs integrate PK and CK \citep{cheng2024understandinginterplayparametriccontextual,su-etal-2024-semi,tao2025lostinthelaterframeworkquantifyingcontextual}, potentially improving contextual grounding. However, it remains unclear what the learned PK--CK interaction dynamics in generating the NLEs are, \textit{necessitating a multi-step analysis of PK--CK interactions across longer NLE sequences.}

Prior work \citep{longpre-etal-2021-entity,xu-etal-2024-knowledge-conflicts,minder2025controllable} has primarily focused on uncovering the single-step generation mechanism -- typically the final answer, and modeled only conflicting PK--CK interaction as a binary choice in a rank-1 subspace, thereby overlooking richer forms such as complementary or supportive knowledge \citep{cheng2024understandinginterplayparametriccontextual}. We thus hypothesize that this \textit{rank-1 subspace is not sufficient to disentangle the individual contributions of PK and CK in all types of knowledge interaction scenarios}. Moreover, they did not characterize the step-by-step PK--CK interaction dynamics over longer sequences, such as NLE generation for different knowledge interactions. To accurately understand the knowledge interaction dynamics during NLE generation, we investigate the following research questions:
\begin{description}[nosep]
    \item [RQ1.] \textit{Is a rank-1 projection subspace enough for disentangling PK and CK contributions in all types of knowledge interaction scenarios?}
    \item [RQ2.] \textit{How do individual PK and CK contributions change over the NLE generation steps for different knowledge interactions?}
    \item [RQ3.] \textit{Can we find reasons for hallucinations based on PK--CK interactions?}
    \item [RQ4.] \textit{How is the CoT mechanism aligned with the knowledge interaction subspace?}
\end{description}
We experiment with four publicly available QA datasets for three open-weight instruction-tuned LLMs. In RQ1 (\S\ref{sec:analysis_rq1}), we find that the rank-1 subspace \citep{minder2025controllable} fails to disentangle individual knowledge contributions across knowledge interactions. For RQ2 (\S\ref{sec:analysis_rq2}), we propose a more accurate rank-2 subspace disentanglement and find that during NLE generation, the model utilizes both knowledge sources with a slight preference for PK; towards the final answer generation, the model aligns closely with the CK direction for PK vs CK conflicting examples; for supportive examples, it aligns more with the PK direction. The learned rank-2 subspace also illustrates PK dominance for sequences with hallucinations (RQ3, \S\ref{sec:rq3_hallucination_alignment}). For RQ4 (\S\ref{sec:rq4_cot_alignment}), we observe that the CoT mechanism balances the PK--CK interaction and often reduces PK alignment. Finally, \S\ref{sec:steering_result} verifies through a subspace intervention that the rank-2 subspace behaviorally disentangles the PK--CK interaction, with PK-axis intervention reducing hallucinations. Overall, this work introduces the first systematic framework for multi-step analysis of knowledge interactions via rank-2 subspace disentanglement, advancing understanding of how LLMs integrate internal and external knowledge.

\section{Related Work}
\label{sec:related_work}

\noindent\textbf{Parametric vs.\ contextual knowledge in LLMs.}
A large body of work studies what language models encode as \emph{parametric knowledge} (PK) and how they integrate \emph{contextual knowledge} (CK) at inference time \citep{petroni-etal-2019-language,jiang-etal-2020-know,roberts-etal-2020-much,hagstrom-etal-2025-reality,hagstrom-etal-2026-cub,marjanovic-etal-2024-dynamicqa,yu-etal-2024-revealing,wangsajaya2026emergencecontextcharacteristicssensitivity,10.1162/COLI.a.621}. Early work showed that factual knowledge can be elicited directly from model parameters \citep{petroni-etal-2019-language}, while probing and representation analyses examined how context and memory are encoded and combined across layers \citep{NEURIPS2020_1457c0d6,tenney-etal-2019-bert,bi2026parameters}. 

More recent studies focus on PK--CK interaction and conflict. \citet{cheng2024understandinginterplayparametriccontextual} analyzed suppression effects between PK and CK under different interaction regimes, while \citet{xu-etal-2024-knowledge-conflicts} categorized conflict types (context--memory, inter-context, intra-memory) and their behavioral consequences. Several approaches aim to improve PK--CK balance through training or inference-time interventions, including enhanced pretraining or fine-tuning \citep{zhang2024evaluatingexternalparametricknowledge}, context-aware representation manipulation \citep{yuan-etal-2025-exploiting}, contrastive decoding \citep{zhao-etal-2024-enhancing}, and lightweight steering that increases context sensitivity without weight updates \citep{wang-etal-2025-continuously}. \citet{hagstrom-etal-2026-cub} empirically compares such context usage manipulation approaches in a joint framework, highlighting the trade-offs between them. Our work complements this line of work by introducing a geometric framework that disentangles PK and CK in a rank-2 subspace and enables multi-step analysis during generation. 

\noindent\textbf{Natural Language Explanations (NLEs) and reasoning.}
NLEs are widely used to expose or guide model reasoning in complex tasks \citep{NEURIPS2018_4c7a167b,atanasova-etal-2020-generating-fact}. Explanation-based supervision improves QA performance \citep{rajani-etal-2019-explain}, and Chain-of-Thought (CoT) prompting elicits stepwise reasoning that enhances multi-hop and arithmetic reasoning \citep{wei2022chain,lampinen-etal-2022-language}. However, multiple studies show that NLEs and CoTs are often unfaithful to the model’s true decision process, functioning as post-hoc rationalizations rather than causal explanations \citep{atanasova-etal-2023-faithfulness,turpin2023language,siegel-etal-2024-probabilities,yuan-etal-2025-graph,wang-atanasova-2025-self,sun2026investigatinginterplaycontextualparametric}. Our work addresses this gap by providing a token-level, representation-based analysis of how PK and CK jointly shape NLE generation, linking explanation faithfulness and hallucination to internal knowledge alignment.

\noindent\textbf{Probing, subspaces, and identifiability.}
Probing methods reveal that linguistic and factual properties are encoded in low-dimensional subspaces of LM representations \citep{hewitt-manning-2019-structural,clark-etal-2019-bert}. The superposition hypothesis further suggests that multiple features are compressed into overlapping linear subspaces \citep{elhage2022superposition}. Building on this view, \citet{minder2025controllable} modeled PK--CK conflict using a single controllable direction (rank-1 subspace). However, when PK and CK are jointly encoded, such a representation is non-identifiable and cannot disentangle individual contributions. We generalize this approach to a rank-2 projection subspace, establishing identifiability and enabling fine-grained, multi-step analysis of PK--CK interaction dynamics during generation.

\section{Method}\label{sec:method}

\subsection{Task Formulation} \label{sec:task_formulation}

Let $\mathcal{Q}$ and $\mathcal{C}$ denote two disjoint sets of queries and contexts, respectively, and consider a QA dataset $\mathcal{D} \subseteq \mathcal{Q} \times \mathcal{C}$. For a question--context pair $(q, c) \in \mathcal{D}$, with $q \in \mathcal{Q}$ and $c \in \mathcal{C}$, a language model $p$ generates both an answer $a$ and an NLE with $n$ tokens as $\mathcal{E} = \{e_i\}_{i=1}^n$. We distinguish between three forms of answer generation: (i) the parametric answer $a(q, \phi)$, produced by recalling PK (independent of $c$, $\phi$ denotes no context), (ii) the contextual answer $a(q, c)$, obtained by leveraging the provided context while disregarding parametric recall, and (iii) the final predicted answer $a$ combining both the provided context and the PK recall guided by the PK--CK knowledge interaction.

During the sequential generation of $\mathcal{E}$, each token $e_i$ is influenced by the interaction between $a(q, \phi)$ and $a(q, c)$. This evolving interaction guides the generation of NLE $\mathcal{E}$, illustrating the decision-making process of the final predicted answer $a$. To analyze this process, we quantify the contribution of parametric knowledge ($\alpha_i^p$) and contextual knowledge ($\alpha_i^c$) at each NLE generation step $i$, and we define their difference as
\begin{equation}
    \Delta_i = {\alpha_i}^p - {\alpha_i}^c, i \in [1, n]
\end{equation}
We normalize individual knowledge contribution $\alpha_i \in [0, 1]$ at each NLE sequence step to compare the influence of PK and CK at each sequence step ($\Delta_i=0$ indicates balance between PK and CK, $\Delta_i=1$ indicates PK dominance, and $\Delta_i=-1$ indicates CK dominance) and how this interaction shifts throughout the NLE generation. By tracking $\Delta_i$ across all NLE generation steps, we aim to characterize the interaction and the shifting balance between parametric and contextual sources throughout NLE generation.

\subsection{Identify Different PK--CK Interactions} \label{sec:identify_intearction_types}
To characterize different types of PK--CK interactions guided by individual knowledge contribution, we draw from \citet{minder2025controllable}, who analyze intent-driven answer control, i.e., controlling the model towards a specific answer generation aligned with the intent of following either PK or CK. Let $w_c$ (instruction to follow only CK) and $w_p$ (instruction to follow only PK) denote intents toward predicting $a(q, c)$ and $a(q, \phi)$, respectively, for a given $(q, c)$.  

\citet{minder2025controllable} restrict intent to conflicting cases, where $a(q, c) \neq a(q, \phi)$ and the final answer $a$ is determined by either $w_c$ or $w_p$. We extend their conflict-only framing, following \citet{cheng2024understandinginterplayparametriccontextual}, to encompass a broader set of interactions:
\begin{itemize}[nosep]
    \item Supportive: $a(q, c) = a(q, \phi) = a$; PK and CK reinforce the same outcome.
    \item Complementary: $a(q, c) \neq a(q, \phi)$ but $a = a(q, c) \cup a(q, \phi)$; neither source is sufficient alone, but their combined information yields the final answer.
    \item Conflicting: $a(q, c) \neq a(q, \phi)$ and $(a = a(q, c)) \oplus (a = a(q, \phi))$; the final answer aligns exclusively with one knowledge source while rejecting the other.
    \item Irrelevant: $c \perp q$ such that $a(q, c)$ is null or nonsensical; the context $c$ contains no semantic information relevant to the query $q$. In this case, the model ignores the context entirely and relies solely on PK ($a = a(q, \phi)$) to formulate a response.
    \item Knowledge Suppression (noise): $a(q, c) = a(q, \phi) \neq a$; both sources agree on an answer, yet the model fails to utilize either, producing an erroneous or unrelated prediction.  
\end{itemize}

To represent how the model balances contextual ($w_c$) and parametric ($w_p$) influences, we extend this intent-driven answer control, and $w_b$ denotes the joint intent of $w_c$ and $w_p$ that illustrates the intrinsic model behavior in generating the final answer $a$, considering both the PK and CK governed by the underlying interaction scenario. 
Unlike in \citet{minder2025controllable}, where $w_c$ and $w_p$ are represented using one orthonormal direction in a learned rank-1 projection subspace, we represent $w_c$ and $w_p$ using two orthonormal directions, and $w_b\rightarrow w_c$ and $w_b\rightarrow w_p$ indicate the individual contributions of CK and PK, respectively, in generating $a$. Then the joint intent $w_b = f(w_c, w_p)$ indicates different knowledge interaction scenarios decided by individual contributions from $w_c$ and $w_p$. We learn this knowledge interaction function using a rank-2 projection subspace.

\subsection{Localize Intent-Guided PK--CK Interactions}\label{sec:intent_encoding}
In this section, we describe where the intent-guided knowledge interaction emerges in the model space, followed by how the LLM encodes it.

\noindent \textbf{Identifying important layers for rank-1 projection subspace.} (showing the individual knowledge contribution towards generating either $a(q,c)$ or $a(q,\phi)$, i.e, $w_c$ or  $w_p$ respectively). 
We follow the activation patching-based mechanistic interpretability approach, Patchscope \cite{pmlr-v235-ghandeharioun24a}. \citet{minder2025controllable} utilizes a Patchscope-based iterative search over all layers to find the best layer corresponding to the maximum likelihood of the PK and CK answers. They construct two minimally different prompts $s$ and $t$ as `source' and `target', with the same $q$ and $c$, only differing by the intent $w_c$ or $w_p$, resulting in two different intended answers. 
To identify the layers capturing the intent present in $s$, during the forward pass of the target $p(.|t)$, the activation from the hidden state of a particular layer is replaced by the activations from the same layer with the source $p(.|s)$, resulting in $\tilde{p}(.|t) \approx p(.|s)$. 
Since they only consider the `conflicting' behaviors in understanding the knowledge interaction in LLMs, they construct two patching datasets capturing opposite directions: $\mathcal{D}_w^{(c \rightarrow p)} = \{s:((q,c,w_c), a(q,c)), \; t:((q,c,w_p), a(q,\phi))\}$ for encoding the CK direction, and $\mathcal{D}_w^{(p \rightarrow c)} = \{s:((q,c,w_p), a(q,\phi)), \; t:((q,c,w_c), a(q,c))\}$ for encoding the PK direction. Layers capturing CK and PK directions, respectively, are selected as:
\begin{equation}
\label{eq:Lsets_r1}
\begin{aligned}
\mathbb{L}_{c \rightarrow p} &= \{\, l \in [1,L] \mid \tilde{p}(\ac \mid q,c,w_p) \ge \tau_c \,\},\\
\mathbb{L}_{p \rightarrow c} &= \{\, l \in [1,L] \mid \tilde{p}(\ap \mid q,c,w_c) \ge \tau_p \,\},
\end{aligned}
\end{equation}
\noindent where $\tau_c$, $\tau_p$ are hyperparameters. For more details, please refer to \citet{minder2025controllable}.

\noindent \textbf{Encoding the intent via rank-1 subspace projection.} Once we identify the important layers capturing the binary interaction between PK and CK for conflicting behavior, we aim to identify a subset of model weights where the LLM encodes this binary interaction. \citet{minder2025controllable} hypothesize that LLMs encode knowledge interaction within their parameter space using a low-rank projection subspace. They model this projection subspace $\mathbf{P} \in \mathbb{R}^{d \times d}$ ($d$ is the model hidden representation size) by a rank-1 unit norm direction $\vec{\mathbf{u}} \in \mathbb{R}^d$ indicating the unidirectional PK--CK conflicting interaction as $\mathbf{P}=\vec{\mathbf{u}}\vec{\mathbf{u}}^T$. At any sequence step $i$, the hidden representation $\vec{\mathbf{h_i}} \in \mathbb{R}^d$ for the token $e_i$ in the NLE $\mathcal{E}$ can be linearly decomposed as:
\begin{align}
\vec{\mathbf{h_i}} &= \mathbf{(I - P)}\vec{\mathbf{h_i}} + \vec{\mathbf{u}}\,\langle \vec{\mathbf{u}}^T, \vec{\mathbf{h_i}} \rangle
\end{align}
    $\mathbf{(I - P)}$ component of $\vec{\mathbf{h_i}}$ captures other properties in the embedding space, and the $\langle \vec{\mathbf{u}}^T, \vec{\mathbf{h_i}} \rangle$ captures the PK--CK knowledge contribution in the PK--CK conflicting direction $\vec{\mathbf{u}}$. 
We hypothesize that the rank-1 projection subspace $\mathbf{P}$ 
cannot disentangle the individual knowledge contributions across different knowledge interaction scenarios. Also, we theoretically argue that this $\mathbf{P}$ fails to satisfy `bijective' properties of mapping from knowledge direction to individual knowledge contribution for all types of knowledge interaction in \cref{thm:nonidentifiability} of \S\ref{app:additional_results}. Hence, we propose to model the PK--CK interaction for different knowledge interactions using a rank-2 projection subspace.

\noindent \textbf{Identifying important layers for rank-2 projection subspace} (showing the individual knowledge contribution towards generating $a$, i.e, $w_b \rightarrow w_c$ and $w_b \rightarrow w_p$). We again follow Patchscope
and prepare two patching datasets $\mathcal{D}_w^{(b \rightarrow p)} = \{s:((q,c,w_b), a), \; t:((q,c,w_p), a(q,\phi))\}$ for encoding the contribution of PK interacting with CK in generating $a$, and $\mathcal{D}_w^{(b \rightarrow c)} = \{s:((q,c,w_b), a), \; t:((q,c,w_c), a(q,c))\}$ for encoding the contribution of CK interacting with PK in generating $a$. Important layers $\mathbb{L}_{b \rightarrow c}$ and $\mathbb{L}_{b \rightarrow p}$ are selected as:
\begin{equation}
\label{eq:Lsets_r2}
\begin{aligned}
\mathbb{L}_{b \rightarrow p} &= \{\, l \in [1,L] \mid \tilde{p}(a \mid q,c,w_p) \ge \tau_p \,\},\\
\mathbb{L}_{b \rightarrow c} &= \{\, l \in [1,L] \mid \tilde{p}(a \mid q,c,w_c) \ge \tau_c \,\},
\end{aligned}
\end{equation}
\noindent where $\tau_p$, $\tau_c$ are hyperparameters (details are described in \cref{tab:app_patching_hyperparameters} in \S\ref{app:hyperparameters}).

\noindent \textbf{Encoding the intent via rank-2 subspace projection.}\label{sec:u_learning_optimization} Once we identify important layers $\mathbb{L}_{b \rightarrow c}$, and $\mathbb{L}_{b \rightarrow p}$ for $w_b \rightarrow w_c$ and $w_b \rightarrow w_p$ respectively, we learn the joint intent function $f$ encoding how individual knowledge contributions $w_b \rightarrow w_c$ and $w_b \rightarrow w_p$ are mixed to generate the final answer $a$ using the common layers from $\mathbb{L}_{b \rightarrow c}$ and $\mathbb{L}_{b \rightarrow p}$ via a rank-2 projection subspace $\mathbf{P}$ spanned by two orthogonal directions $\vec{\mathbf{u}} \in \mathbb{R}^{d \times 2}$ as $\mathbf{P}=\vec{\mathbf{u}}(\vec{\mathbf{u}}^T\vec{\mathbf{u}})^{-1}\vec{\mathbf{u}}^T$. Since we aim to identify individual contributions from the individual directions, following \citet{minder2025controllable}, we consider those two basis vectors as orthonormal, and hence the rank-2 projection subspace $\mathbf{P}$ is reduced to $\mathbf{P}=\vec{\mathbf{u}}\vec{\mathbf{u}}^T$, since $\vec{\mathbf{u}}^T\vec{\mathbf{u}} = \mathbf{I_2}$. Considering $\vec{\mathbf{u}} = [\vec{\mathbf{u_c}},\vec{\mathbf{u_p}}]$, at any sequence step $i$, the hidden representation $\vec{\mathbf{h_i}} \in \mathbb{R}^d$ for the token $e_i$ in the NLE $\mathcal{E}$ can be linearly decomposed as:
\begin{align}
\vec{\mathbf{h_i}} &= \mathbf{(I - P)}\vec{\mathbf{h_i}} + \vec{\mathbf{u_c}}\,\langle \vec{\mathbf{u_c}}^T, \vec{\mathbf{h_i}} \rangle + \vec{\mathbf{u_p}}\,\langle \vec{\mathbf{u_p}}^T, \vec{\mathbf{h_i}} \rangle
\end{align}
\noindent \textbf{Learning the rank-2 projection subspace.} Let $\tilde{\vec{\mathbf{h}}}^{w_p}
= (\mathbf{I}-\mathbf{P})\,\vec{\mathbf{h}}
+ \vec{\mathbf{u}}_c\,\langle \vec{\mathbf{u}}_c^{T}, \vec{\mathbf{h}} \rangle
+ \vec{\mathbf{u}}_p\,\langle \vec{\mathbf{u}}_p^{T}, \vec{\mathbf{h}}^{\,p} \rangle$,
and $\tilde{\vec{\mathbf{h}}}^{w_c}
= (\mathbf{I}-\mathbf{P})\,\vec{\mathbf{h}}
+ \vec{\mathbf{u}}_c\,\langle \vec{\mathbf{u}}_c^{T}, \vec{\mathbf{h}}^{\,c} \rangle
+ \vec{\mathbf{u}}_p\,\langle \vec{\mathbf{u}}_p^{T}, \vec{\mathbf{h}} \rangle$
be two patched decompositions of the last-token hidden representation of the NLE $\mathcal{E}$ by replacing the activation from a particular layer
$l \in \mathbb{L}_{b \rightarrow p} \cup \mathbb{L}_{b \rightarrow c}$
during the forward pass of $p(\cdot \mid q, c, w_b)$ with the activations from the same layer of the forward pass of
$p(\cdot \mid q, c, w_p)$ and $p(\cdot \mid q, c, w_c)$, respectively.
Then we learn the projection subspace $\mathbf{P}$ spanned by the two orthonormal basis vectors
$\vec{\mathbf{u}} = [\vec{\mathbf{u}}_c,\vec{\mathbf{u}}_p]$
by minimizing the two objectives sequentially, maintaining orthonormal directions between
$\vec{\mathbf{u}}_c$ and $\vec{\mathbf{u}}_p$\footnote{We initialize the projection weight from two orthogonal vectors and \texttt{nn.utils.parametrizations.orthogonal()} automatically maintains orthogonality after each optimization step.}:
\begin{align}
\mathcal{J}_p
&= -\frac{1}{N} \sum_{n=1}^{N}
\log \bigl( \tilde{p}\!\left( a(q_n, \phi)
\mid \tilde{\vec{\mathbf{h}}}_{(n)}^{w_p} \right) \bigr), \\
\mathcal{J}_c
&= -\frac{1}{N} \sum_{n=1}^{N}
\log \bigl( \tilde{p}\!\left( a(q_n, c_n)
\mid \tilde{\vec{\mathbf{h}}}_{(n)}^{w_c} \right) \bigr).
\end{align}

The methodology of assigning the PK--CK direction to the basis vector $\vec{\mathbf{u}}$ as a validation step is described in \S\ref{app:pk_ck_direction_assignment}.
\paragraph{Normalized PK--CK score.}\label{sec:pk_ck_contrib_norm} At any sequence step $i$, given the hidden representation $\vec{\mathbf{h_i}} \in \mathbb{R}^d$ for the token $e_i$ in the NLE $\mathcal{E}$, let $c_i = \lvert\langle \vec{\mathbf{u_c}}^T, \vec{\mathbf{h_i}} \rangle\lvert$ and $p_i = \lvert\langle \vec{\mathbf{u_p}}^T, \vec{\mathbf{h_i}} \rangle\lvert$, then the normalized contribution from CK and PK can be computed as ${\alpha_i}^c = c_i / (c_i + p_i)$ and ${\alpha_i}^p = p_i / (c_i + p_i)$, and ${\alpha_i}^c, {\alpha_i}^p \in [0, 1]$ satisfying the identifiability of individual knowledge contribution (\cref{thm:nonidentifiability}) under rank-2 subspace.

\section{Results}
\label{sec:results}

\subsection{Experimental Setting} \label{sec:exp_setting}

\textbf{Dataset and Model:} We conduct experiments on three open-weight decoder-only instruct-based LLMs: Llama-3.1-8B \citep{grattafiori2024llama3herdmodels}, Gemma-2 9B \citep{gemmateam2024gemma2improvingopen} and Mistral-v0.3 7B \citep{jiang2023mistral7b} using four publicly available QA datasets: BaseFakepedia, MultihopFakepedia \cite{minder2025controllable}, StrategyQA \cite{geva-etal-2021-aristotle}, and OpenBookQA \cite{mihaylov-etal-2018-suit,cheng2024understandinginterplayparametriccontextual}. BaseFakepedia is a knowledge conflict dataset containing queries with 23 relations from Wikipedia. MultihopFakepedia is an extension of BaseFakepedia that contains queries that require extra-hop reasoning to generate answers. Both BaseFakepedia and MultihopFakepedia datasets are utilized for the rank-1 subspace analysis in \citet{minder2025controllable}. StrategyQA is a multi-hop reasoning QA dataset that includes implicit reasoning steps in the queries. OpenBookQA is a commonsense reasoning-based QA dataset. Both StrategyQA and OpenBookQA datasets include `supportive', `complementary', `conflicting', `irrelevant', and `knowledge suppression' PK--CK interactions, illustrated in \citet{cheng2024understandinginterplayparametriccontextual}. Prompt templates for intent-driven answer control, NLE and CoT generations are described in \S\ref{app:prompt_template}.

\subsection{RQ1: Is a Rank-1 Projection Subspace Enough for Disentangling PK and CK Contribution Across All Types of Knowledge Interaction Scenarios?}\label{sec:analysis_rq1}
\begin{figure}[t]
    \centering
    \includegraphics[width=1.0\linewidth]{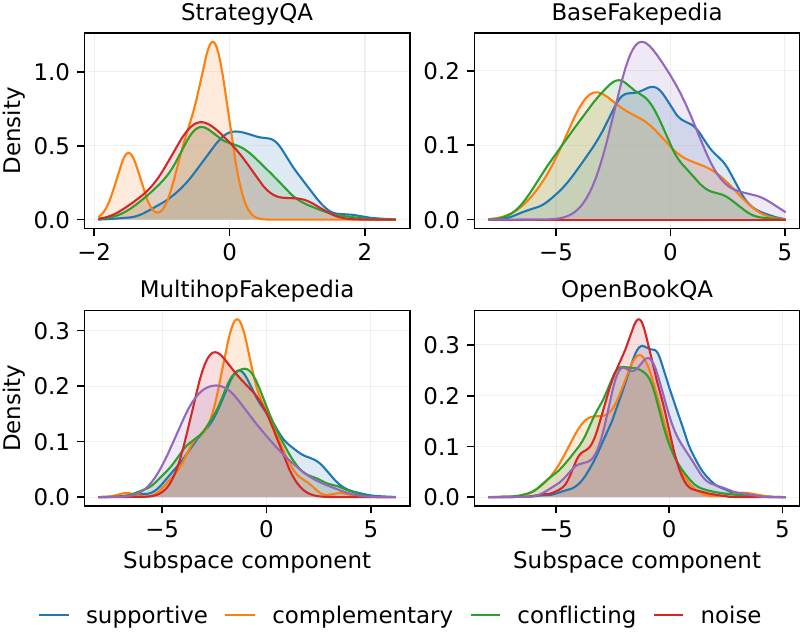}
    \caption{Kernel Density Estimate (KDE) of the PK--CK subspace component $\langle \vec{\mathbf{u}}^{T}, \vec{\mathbf{h}}_{i} \rangle$ across different knowledge interaction types for four QA datasets using Llama-3.1-8B-Instruct model. See similar results for the other two models in \cref{app:additional_results}.}
    \label{fig:rank1_proj_contribution_llama}
\end{figure}

\begin{figure}[t]
    \centering
    \includegraphics[width=\linewidth]{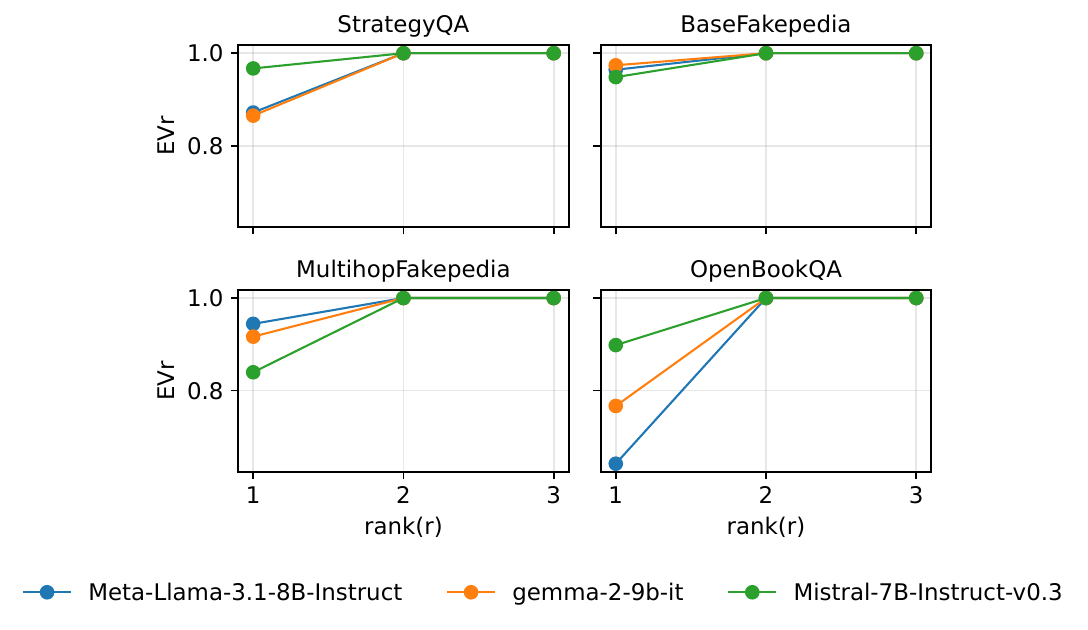}
    \caption{Cumulative explained variance ($EV_r$) at rank(r) from the three models using the four QA datasets. At rank-2, it reaches ~$1.0$ value, indicating sufficiency in capturing different knowledge interaction variants.}
    \label{fig:explained_variance_vs_rank}
\end{figure}

We verify the sufficiency of rank-1 subspace in disentangling PK--CK contribution in different types of knowledge interactions using two metrics: (a) \textit{Subspace component} \citep{minder2025controllable}, which indicates the strength of the hidden representation of answer tokens along the direction of the basis vector $\vec{\mathbf{u}}$ in the rank-1 subspace. If a rank-1 subspace is sufficient in disentangling PK--CK contributions, then the distribution of subspace components of answer tokens for different interactions should not overlap and have a mean far from zero, indicating a small contribution in the orthogonal $\mathbf{(I - P)}$ direction; (b) \textit{Explained Variance (Cumulative)} \citep{Wall2003,JMLR:v26:24-0699}, which indicates the minimum rank $r$ required to explain the variance in data distributions for different types of knowledge interactions. It measures the strength of top-$r$ singular values among all possible singular values present in the data distribution, and at rank-$r$, the fraction should be closer to $1.0$ if it is sufficient to explain. More details about the metrics are discussed in \cref{app:eval_metric}. \cref{fig:rank1_proj_contribution_llama} shows the \textit{subspace component} distribution of answer tokens for different knowledge interaction types overlap, and their means converge to $0$, indicating insufficiency of rank-1 subspace. \cref{fig:explained_variance_vs_rank} also verifies rank-2 subspace as the minimum projection subspace to capture data variance and disentangle PK--CK contribution.

\subsection{RQ2: How Do Individual PK and CK Contributions Change Over the NLE Generation for Different Knowledge Interactions?}\label{sec:analysis_rq2}
\begin{figure}[t]
  \centering
  \begin{subfigure}{\linewidth}
    \centering
    \includegraphics[width=\linewidth]{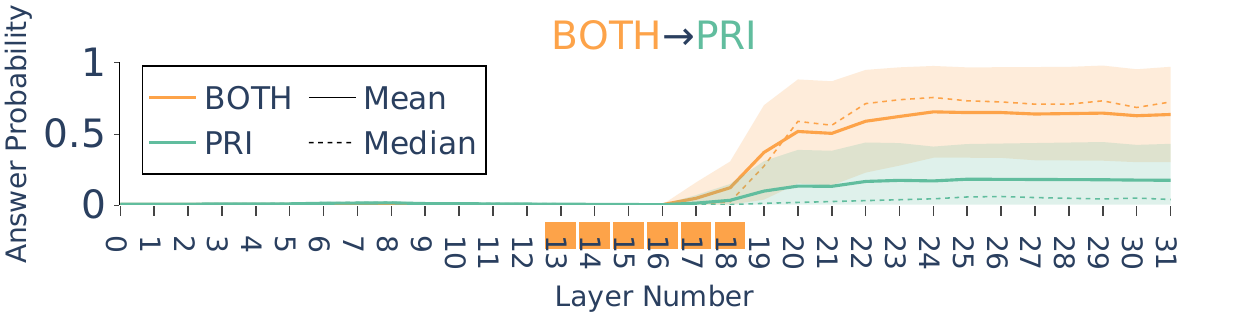}
    \caption{$\mathcal{D}_w^{(b \rightarrow p)}$}
    \label{fig:patchscope_bp}
  \end{subfigure}\hfill
  \begin{subfigure}{\linewidth}
    \centering
    \includegraphics[width=\linewidth]{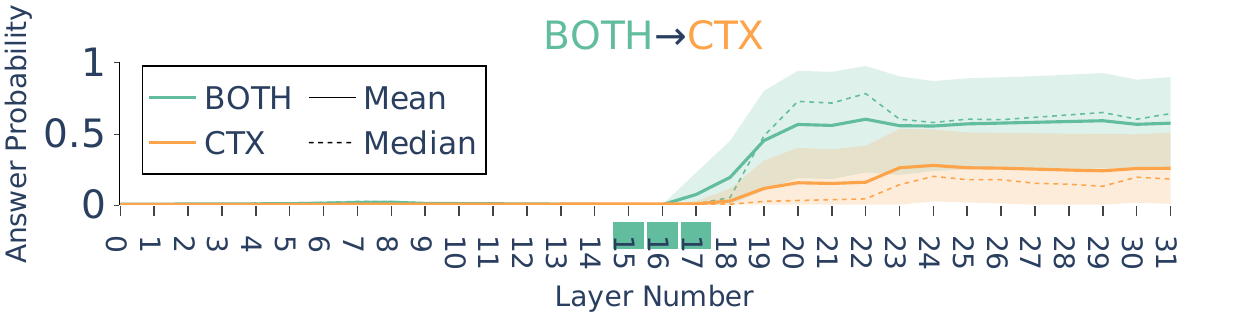}
    \caption{$\mathcal{D}_w^{(b \rightarrow c)}$}
    \label{fig:patchscope_bc}
  \end{subfigure}
  \caption{Patchscope on OpenBookQA from Llama-3.1-8B-Instruct. PRI indicates PK, and CTX indicates CK. (a) Activation patching on $\mathcal{D}_w^{(b \rightarrow p)}$. (b) Activation patching on $\mathcal{D}_w^{(b \rightarrow c)}$. Computation details of patchscope are described in \cref{app:compute_patchscope}. }
  \label{fig:patching_layer_openbookqa_llama3.1_8b}
\end{figure}

We investigate the dynamics of individual PK and CK contribution ${\alpha_i}^p$  and ${\alpha_i}^c$ at every sequence step $i$ over the NLE $\mathcal{E}$ generation. Among the four datasets, OpenBookQA shows the largest jump in cumulative explained variance from rank-1 to rank-2 across all models (\cref{fig:explained_variance_vs_rank}). We therefore use OpenBookQA as our development set to (i) locate intent-sensitive layers via activation patching and (ii) learn the rank-2 projection subspace. We then freeze these choices and evaluate the resulting rank-2 subspace on all datasets. 
StrategyQA yields nearly identical subspaces (directional similarity $>0.9$ on both axes; see §\ref{app:robustness}).
We first localize the layers encoding individual PK and CK contributions.
\cref{fig:patching_layer_openbookqa_llama3.1_8b} shows that patching from $w_b \rightarrow w_p$ (BOTH→PRI) (L13–18) yields a larger probability gain than $w_b \rightarrow w_c$ (BOTH→CTX) (L15–17), indicating that Llama-3.1-8B-Instruct relies more on PK than on CK for OpenBookQA. We observe similar results for gemma-2-9b-it and Mistral-7B-Instruct-v0.3 ( \cref{fig:patching_layer_openbookqa_gemma_2_9b_it,fig:patching_layer_openbookqa_mistral_7b_instruct_v0.3}). This PK dominance in a context-rich task suggests that the model tends to recall commonsense information from PK rather than grounding answers in CK, highlighting potential for future work on context-sensitive knowledge control.

\begin{figure}[t]
    \centering
    \begin{subfigure}{\linewidth}
        \centering
        \includegraphics[width=\linewidth]{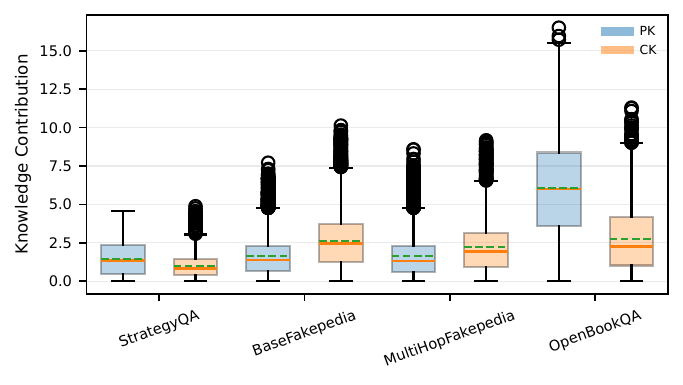}
        \caption{Individual PK--CK contribution $p_i$ and $c_i$ (\S\ref{sec:pk_ck_contrib_norm}) in generating the answer token for Llama-3.1-8B-Instruct.}
        \label{fig:rank2_pk_ck_answer_token}
    \end{subfigure}

    \vspace{0.3em}

    \begin{subfigure}{\linewidth}
        \centering
        \includegraphics[width=1.0\linewidth]{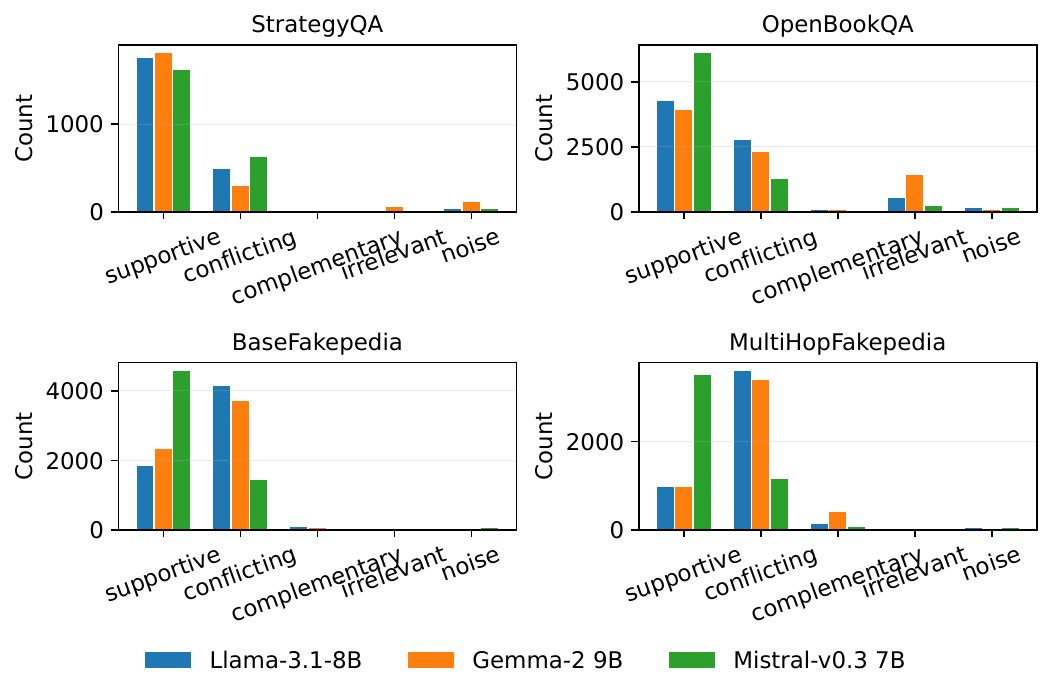}
        \caption{Distribution of knowledge interaction types across datasets.}
        \label{fig:data_distribution}
    \end{subfigure}

    \caption{Answer-level PK--CK alignment and distribution of knowledge interaction types across datasets.}
    \label{fig:answer_token_and_data_distribution}
\end{figure}

\begin{figure}[!ht]
    \centering
    \includegraphics[width=1.0\linewidth]{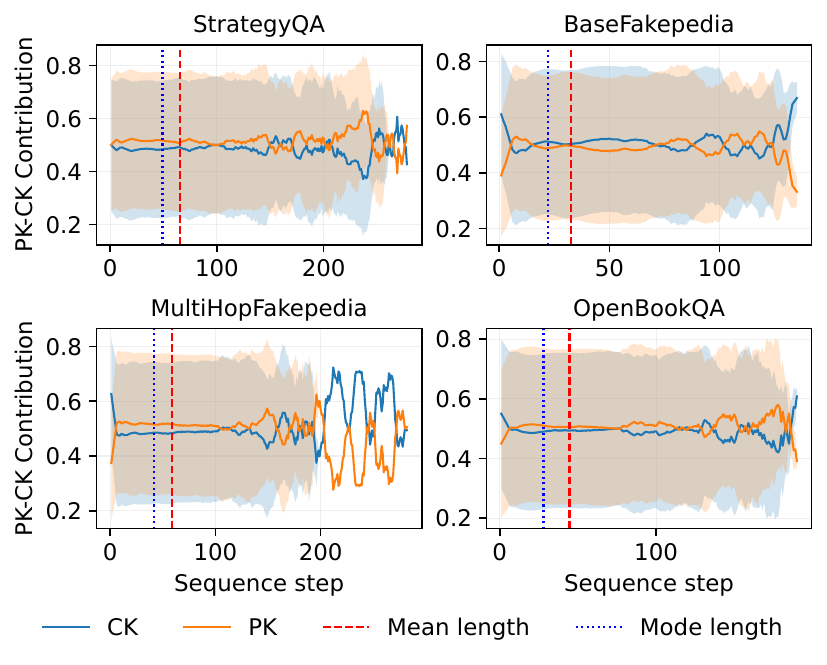}
    \caption{PK--CK interaction dynamics over the sequence steps for Llama-3.1-8B-Instruct. The dotted red and blue lines indicate the mean and mode of NLE lengths. \cref{fig:pkck_dynamics_llama_detailed} contains PK--CK dynamics for different interaction scenarios.}
    \label{fig:pkck_dynamics_llama}
\end{figure}

To identify the individual knowledge contribution in generating the final answer token $a$, we project its hidden representation $\vec{\mathbf{h_a}}$ from the last layer of $l \in \mathbb{L}_{b \rightarrow p} \cup \mathbb{L}_{b \rightarrow c}$ to the rank-2 projection subspace to compute the $p_i$ and $c_i$ as the PK--CK contributions\footnote{We do not normalize for single token answer.}. \cref{fig:rank2_pk_ck_answer_token} indicates an overall higher CK contribution for BaseFakepedia and MultihopFakepedia, and a higher PK contribution for StrategyQA and OpenBookQA for the Llama-3.1-8B-Instruct model. We further investigate the distribution of different knowledge interactions (following \S\ref{sec:identify_intearction_types}) in Fig. \ref{fig:data_distribution} and find that both BaseFakepedia and MultihopFakepedia contain more conflicting examples than other knowledge interaction types (\S\ref{sec:identify_intearction_types}). 
Our findings match both dataset composition and prior findings: Fakepedia variants often contain \textit{examples with conflicting CK when models are found to suppress PK}; StrategyQA/OpenBookQA have examples that rely on common sense priors and have supportive (i.e., non-conflicting) contexts when models default to parametric recall \citep{cheng2024understandinginterplayparametriccontextual,tao-etal-2024-context}. Overall, we conclude that \textit{for conflicting examples, the model aligns more with the CK direction and for supportive examples, the model aligns more with the PK direction in the rank-2 projection subspace}. 

To understand the knowledge interaction dynamics during NLE generation,
We analyze variation in ${\alpha_i}^p$ and ${\alpha_i}^c$ over all sequence steps of NLE generation for different knowledge interaction scenarios. \cref{fig:pkck_dynamics_llama} shows that for all datasets, \textit{during most of the NLE generations,} 
\textit{the model starts with a higher CK, then considers both PK and CK with slight prioritization of PK.} However, for longer NLEs, CK and PK compete with each other with higher fluctuation. Longer NLEs indicate difficult examples with higher depth in multi-hop reasoning and higher token uncertainty (from \cref{fig:entropy_vs_nle_quartile}), which force the model to iteratively reconcile PK with CK, resulting in this fluctuating behavior.

\begin{figure}[t]
    \centering\includegraphics[width=\linewidth]{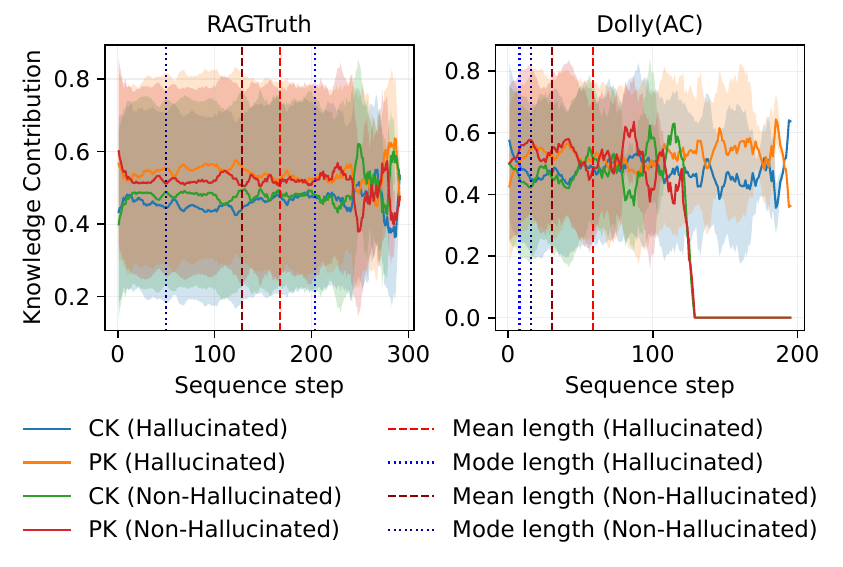}
    \caption{PK--CK interaction dynamics over the sequence step from Llama-3.1-8B-Instruct for the two RAG hallucination datasets.}
    \label{fig:rq3_hal_pk_ck_dynamics}
\end{figure}

\begin{figure}[t]
    \centering
    \includegraphics[width=\linewidth]{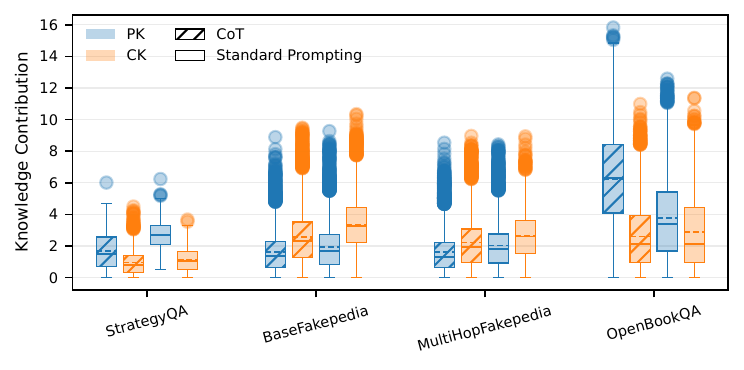}
    \caption{Comparison in individual PK--CK contribution $p_i$ and $c_i$ (\S\ref{sec:pk_ck_contrib_norm}) in generating the answer token $a$ for all the datasets from Llama-3.1-8B-Instruct model between CoT and standard prompting. }
    \label{fig:cot_vs_standard_prompt_pk_ck_answer_token}
\end{figure}

\subsection{RQ3: Can We Find Reasons for Hallucinations Based on PK--CK Interactions?}\label{sec:rq3_hallucination_alignment}

To characterize the knowledge interaction dynamics during the NLE generation in terms of context faithfulness, we compare hallucinated vs non-hallucinated responses in the rank-2 projection subspace. Following \citep{sun2025redeep}, we utilize two commonly used RAG hallucination datasets: RAGTruth \cite{niu-etal-2024-ragtruth} and Dolly (AC) \cite{hu-etal-2024-knowledge}, of sizes $18240$ and $297$ examples, respectively, containing examples from QA, Summarisation and Information Extraction. Each dataset provides high-quality human-annotated spans of hallucinated content across responses from multiple models. Importantly, whether a span is labelled as hallucinated is \emph{model-dependent}: the same RAG input may yield hallucinated text for one model but not for another due to differences in the model's PK. Due to the limited number of models covered in the two datasets, we find hallucination labels only for the Llama-3.1-8B-Instruct model in these two datasets. \cref{fig:rq3_hal_pk_ck_dynamics} illustrates the PK--CK knowledge interaction dynamics. The gap between PK and CK is much higher for the examples with hallucinated spans than for the other examples across the sequence steps. This result also aligns with similar observations of \textit{positive correlation of PK and hallucination} in \citet{sun2025redeep}.

\subsection{RQ4: How is the CoT mechanism aligned with the knowledge interaction subspace?}\label{sec:rq4_cot_alignment}

To verify whether reasoning-based prompting CoT helps the model to stay aligned with the CK and reduces reliance on PK, we compare the PK--CK contribution in generating the final answer between standard prompting and CoT prompting.   
Figure \ref{fig:cot_vs_standard_prompt_pk_ck_answer_token} indicates that \textit{CoT maintains similar CK alignment compared to standard prompting for all the datasets, and also reduces PK alignment} except for the OpenBookQA dataset.

\section{Steering Model Behavior via Projection Subspace Intervention}\label{sec:steering_result}
Instead of using the explicit intent instruction $w_p$ (follow prior knowledge only), $w_c$ (follow context knowledge only) and $w_b$ (consider both knowledge sources), we steer the hidden representation of the answer token using the equation:
\begin{equation}\label{eq:steering_eq}
    \tilde{\vec{\mathbf{h_i}}} = \mathbf{(I - P)}\vec{\mathbf{h_i}} + \vec{\mathbf{u}}_c\, c_i + \vec{\mathbf{u}}_p\, p_i
\end{equation}
where $c_i, p_i \in \mathbb{R}^+$ are weights along the CK and PK direction, respectively.
We perform this steering in three settings: $p_i  > c_i$ (similar to $w_p$), $p_i < c_i$ (similar to $w_c$), and $p_i = c_i$ (similar to $w_b$) on all four datasets and three models using the learned rank-2 projection subspace and measure accuracy on how well the model predicts the same answer without the explicit intent.

\begin{table}[t]
\centering
\small
\setlength{\tabcolsep}{4pt}
\begin{tabular}{llccc}
\toprule
\textbf{Dataset} & \textbf{Model} & $p_i {>} c_i$ & $p_i {<} c_i$ & $p_i {=} c_i$ \\
\midrule
\multirow{3}{*}{\shortstack[l]{Strategy\\QA}}
  & Llama   & 0.910 & 0.940 & 0.935 \\
  & Gemma   & 0.890 & 0.910 & 0.925 \\
  & Mistral & 0.840 & 0.875 & 0.910 \\
\midrule
\multirow{3}{*}{\shortstack[l]{Base\\Fakepedia}}
  & Llama   & 0.865 & 0.930 & 0.915 \\
  & Gemma   & 0.830 & 0.905 & 0.900 \\
  & Mistral & 0.835 & 0.870 & 0.860 \\
\midrule
\multirow{3}{*}{\shortstack[l]{Multihop\\Fakepedia}}
  & Llama   & 0.870 & 0.900 & 0.915 \\
  & Gemma   & 0.845 & 0.870 & 0.870 \\
  & Mistral & 0.827 & 0.860 & 0.855 \\
\midrule
\multirow{3}{*}{\shortstack[l]{OpenBook\\QA}}
  & Llama   & 0.940 & 0.960 & 0.960 \\
  & Gemma   & 0.917 & 0.925 & 0.920 \\
  & Mistral & 0.890 & 0.903 & 0.925 \\
\bottomrule
\end{tabular}
\caption{Model behavior performance between the answer prediction by intervening in the learned rank-2 subspace and the answer predicted with explicit intent. $p_i {>} c_i$: $p_i{=}2.50, c_i{=}1.95$; $p_i {<} c_i$: $p_i{=}2.0, c_i{=}2.50$; $p_i {=} c_i$: $p_i{=}c_i{=}2.50$.}
\label{tab:robustness_pk_ck_ratio}
\end{table}

\begin{table}[t]
\centering
\small
\setlength{\tabcolsep}{4pt}
\begin{tabular}{lc}
\toprule
\textbf{Steering Condition} & \textbf{Hallucinated / Total} \\
\midrule
No steering & 243 / 450 \\
$p_i = 2.1$ (PK-downweight) & 202 / 450 ($<$0.0001) \\
$c_i = 1.94$ (CK-upweight) & 221 / 450 (0.048) \\
$p_i = 2.1$, $c_i = 1.94$ (Both) & 193 / 450 ($<$0.0001) \\
\bottomrule
\end{tabular}
\caption{Reduction in hallucinated generation by intervening in the learned rank-2 subspace (p-values from Fisher's exact test in the bracket).}
\label{tab:steering_results_hal}
\end{table}

From \cref{tab:robustness_pk_ck_ratio}, the high alignment between steering in the projection subspace and the intent variable that guides the knowledge interaction indeed verifies that the learned projection subspace behaviorally disentangles PK--CK interaction.

For the RAGTruth dataset from Llama-3.1-8B-Instruct, we observe average CK and PK scores of 0.46 ($c_i = 1.8559$) and 0.54 ($p_i = 2.2277$) for hallucinated examples and 0.49 ($c_i = 1.9399$) and 0.51 ($p_i = 2.0927$) for non-hallucinated examples. Using \cref{eq:steering_eq} to obtain the hidden representation at each sequence step, following \citep{sun2025redeep}, we steer 450 examples and regenerate their responses, keeping the generation parameters unchanged. We then compute the truthfulness of the generated response using Llama-3.1-70B-Instruct with the same prompt template as in \citep{sun2025redeep}, and report the number of examples identified as hallucinated in \cref{tab:steering_results_hal}. We compare our hallucination mitigation performance with \citet{sun2025redeep}, which achieves $172$ wins, $179$ ties, and $99$ losses over the non-steered model, resulting in a reduction of $(73/450)$ compared to our reported $(50/450)$ reduction in hallucination generation. However, our result remains statistically significant, indicating a stable reduction in hallucinations compared to the non-steered method. We also note that \citet{sun2025redeep} computes the per-token residual contribution of the attention and the MLP block and hallucination likelihood based on the positive correlation of the MLP contribution and the negative correlation of the attention contribution with the per-token hallucination gold label. \textit{Overall, our work establishes a causal study of hallucination likelihood with PK--CK interaction via PK directional steering-based truthful generation, whereas \citet{sun2025redeep} utilizes all possible concepts relevant for hallucination via residual contribution to detect and mitigate hallucination, resulting in higher but unstable mitigation performance}. However, hallucination mitigation performance in \citep{sun2025redeep} suggests that the PK axis in the learned rank-2 subspace is neither a unique nor a complete causal mechanism for hallucination. This hallucination mechanism can be explained by another projection subspace apart from the PK--CK subspace. But the high alignment of the PK axis indicates that another subspace should not be orthogonal to the PK direction of the learned rank-2 subspace.

\section{Conclusion}
\label{sec:conclusion}
This work shows that the interaction between Parametric Knowledge (PK) and Context Knowledge (CK) in LLMs is inherently multidimensional rather than a binary choice. We introduce a rank-2 projection framework that (i) resolves the non-identifiability of rank-1 probes in disentangling individual knowledge contributions, (ii) disentangles token-level PK and CK components, and (iii) enables the first systematic multi-step analysis of PK--CK dynamics during Natural Language Explanation (NLE) generation. Across four QA datasets and three open-weight instruction-tuned LMs, rank-1 subspaces fail to capture diverse knowledge interaction scenarios, whereas rank-2 suffices to explain the observed variance. Our step-wise analysis further reveals consistent patterns: hallucinated generations align strongly with the PK axis, while context-faithful generations maintain a more balanced PK--CK alignment, and Chain-of-Thought prompting shifts generations toward CK by reducing PK reliance. Intervention in the learned rank-2 subspace enables controlling model behavior towards the desired intent and reduces hallucination generation, thereby verifying the behavioral knowledge disentanglement in that subspace. Beyond NLE generation, this subspace-based analysis provides a general, model-internal signal for studying how LLMs balance internal recall and external grounding.

\section*{Limitations}
\label{sec:limitations}

\noindent \paragraph{\textbf{Orthonormal assumption of PK--CK directions:}} We restrict the subspace to span two orthonormal directions to obtain a pure directional contribution in the LLM knowledge interaction. In principle, the subspace can be spanned by two non-orthogonal directions, but mapping the PK--CK contribution to those directions would be complicated. We leave it for future work. We verify that these two orthonormal directions correspond to the PK and CK knowledge from the steering-based intervention experiments in \cref{sec:steering_result}. 

\noindent \paragraph{\textbf{Rank-3 or higher rank subspace projection in disentangling PK--CK contribution:}} We limit our methodology to rank-2 based on the observed cumulative explained variance across all datasets and models, averaged over subspace rank. We also empirically verify the sufficiency of rank-2 in encoding the knowledge interaction concept. As future work, we can examine the orthogonal component of the subspace to understand the model's sensitivity to it, thereby establishing the need for rank-3. 

\noindent \paragraph{\textbf{Comparing rank-2 subspace with PCA, ICA, or alternative representation steering methods: }}We find standard PCA and ICA-based methods are invalid baselines for disentangling knowledge interactions within the model. PCA finds directions along the maximum variance in the hidden representation matrix of examples and may not correspond to the actual knowledge directions. ICA finds two statistically independent directions in the representation, which are again causally connected to the knowledge directions. PatchScope-based rank-1 or rank-2 subspace learning enforces the learned subspace direction to the knowledge direction. We theoretically show that rank-1 is non-identifiable in disentangling individual PK and CK contributions, and empirically verify the sufficiency of the proposed rank-2 subspace.

\noindent \paragraph{\textbf{Inspecting triggering effects causing the model to shift from CK to PK:}} Inspecting triggering effects that cause the model to shift from CK to PK warrants a different set of experiments, which is not the focus of our work, as we aim to disentangle and model the PK--CK contributions. \citet{sun2025redeep} introduced the external context score (ECS) by calculating the residual contribution from multi-head attention and the parametric knowledge score (PKS) from the FFN blocks’ residual contributions for a particular token, and found a positive correlation between hallucination and PKS. They observe that the model successfully attends to correct context around $65\%$ of the time, but it loses this information over longer sequence generation, leading to lower ECS or higher PKS resulting in hallucination.

\noindent \paragraph{\textbf{Nonlinearity of LLM representations:}} We follow the previous findings on the linear decomposition of several concepts encoded in the token's hidden representation and nonlinearity in the interaction is out of scope for our study.

\noindent \paragraph{\textbf{Evaluation beyond QA:}} We investigate the effectiveness of the rank-2 projection subspace on three open-weight decoder-only instruction-tuned LLMs across four QA datasets due to the following reasons: 1) QA datasets are the only available annotated datasets with different types of knowledge interactions, 2) QA tasks are traditionally used to explore interaction between the model’s own parametric knowledge (PK) and the external context knowledge (CK) as they naturally rely on both types of knowledge. Investigating diverse sequence generation datasets, such as summarization, translation, or dialogue, would broaden the generalization of our findings. However, translation requires interaction between context and multi-lingual parametric knowledge, which would require a different type of study of how context in one language is linked with PK in another language in the model, and where we do not expect different types of knowledge interactions. Same with both summarization and dialogue generation, where they heavily rely on external context (external document, conversation history). Also, any information-seeking queries in a dialogue are equivalent to a question-answering task. 





\section*{Ethical Considerations}
\noindent \textbf{Use of AI Assistants.} The authors used Claude API to assist with language polishing and grammatical corrections, clarity and coherence. All technical contributions, including experimental design choices and final decisions, were made by the authors.

\section*{Acknowledgements}
$\begin{array}{l}\includegraphics[width=1cm]{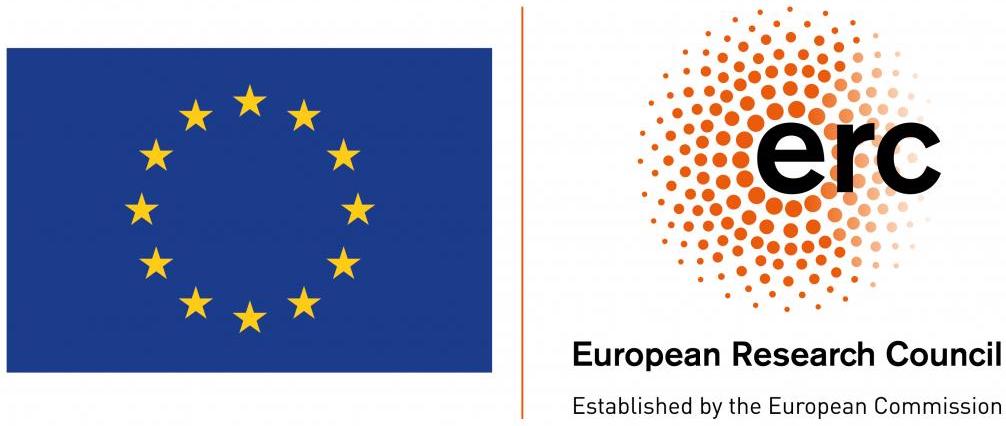} \end{array}$ 
This research was co-funded by the European Union (ERC, ExplainYourself, 101077481) and by VILLUM FONDEN (grant number 40543). Views and opinions expressed are however those of the author(s) only and do not necessarily reflect those of the European Union or the European Research Council. Neither the European Union nor the granting authority can be held responsible for them.





\bibliography{custom}

@inproceedings{zhao-etal-2024-enhancing,
    title = {{Enhancing Contextual Understanding in Large Language Models through Contrastive Decoding}},
    author = "Zhao, Zheng  and
      Monti, Emilio  and
      Lehmann, Jens  and
      Assem, Haytham",
    editor = "Duh, Kevin  and
      Gomez, Helena  and
      Bethard, Steven",
    booktitle = "Proceedings of the 2024 Conference of the North American Chapter of the Association for Computational Linguistics: Human Language Technologies (Volume 1: Long Papers)",
    month = jun,
    year = "2024",
    address = "Mexico City, Mexico",
    publisher = "Association for Computational Linguistics",
    url = "https://aclanthology.org/2024.naacl-long.237/",
    doi = "10.18653/v1/2024.naacl-long.237",
    pages = "4225--4237"
}

@misc{zhang2024evaluatingexternalparametricknowledge,
      title={{Evaluating the External and Parametric Knowledge Fusion of Large Language Models}}, 
      author={Hao Zhang and Yuyang Zhang and Xiaoguang Li and Wenxuan Shi and Haonan Xu and Huanshuo Liu and Yasheng Wang and Lifeng Shang and Qun Liu and Yong Liu and Ruiming Tang},
      year={2024},
      eprint={2405.19010},
      archivePrefix={arXiv},
      primaryClass={cs.CL},
      url={https://arxiv.org/abs/2405.19010}, 
}

@inproceedings{yuan-etal-2025-exploiting,
    title = {{Exploiting Contextual Knowledge in {LLM}s through $\mathcal{V}$-usable Information based Layer Enhancement}},
    author = "Yuan, Xiaowei  and
      Yang, Zhao  and
      Huang, Ziyang  and
      Wang, Yequan  and
      Fan, Siqi  and
      Ju, Yiming  and
      Zhao, Jun  and
      Liu, Kang",
    editor = "Che, Wanxiang  and
      Nabende, Joyce  and
      Shutova, Ekaterina  and
      Pilehvar, Mohammad Taher",
    booktitle = "Proceedings of the 63rd Annual Meeting of the Association for Computational Linguistics (Volume 1: Long Papers)",
    month = jul,
    year = "2025",
    address = "Vienna, Austria",
    publisher = "Association for Computational Linguistics",
    url = "https://aclanthology.org/2025.acl-long.1531/",
    doi = "10.18653/v1/2025.acl-long.1531",
    pages = "31726--31741",
    ISBN = "979-8-89176-251-0"
}

@inproceedings{yuan-etal-2025-graph,
    title = {{Graph-Guided Textual Explanation Generation Framework}},
    author = {Yuan, Shuzhou  and
      Sun, Jingyi  and
      Zhang, Ran  and
      F{\"a}rber, Michael  and
      Eger, Steffen  and
      Atanasova, Pepa  and
      Augenstein, Isabelle},
    editor = "Christodoulopoulos, Christos  and
      Chakraborty, Tanmoy  and
      Rose, Carolyn  and
      Peng, Violet",
    booktitle = "Proceedings of the 2025 Conference on Empirical Methods in Natural Language Processing",
    month = nov,
    year = "2025",
    address = "Suzhou, China",
    publisher = "Association for Computational Linguistics",
    url = "https://aclanthology.org/2025.emnlp-main.1494/",
    doi = "10.18653/v1/2025.emnlp-main.1494",
    pages = "29362--29386",
    ISBN = "979-8-89176-332-6"
}

@inproceedings{yu-etal-2024-revealing,
    title = {{Revealing the Parametric Knowledge of Language Models: A Unified Framework for Attribution Methods}},
    author = "Yu, Haeun  and
      Atanasova, Pepa  and
      Augenstein, Isabelle",
    editor = "Ku, Lun-Wei  and
      Martins, Andre  and
      Srikumar, Vivek",
    booktitle = "Proceedings of the 62nd Annual Meeting of the Association for Computational Linguistics (Volume 1: Long Papers)",
    month = aug,
    year = "2024",
    address = "Bangkok, Thailand",
    publisher = "Association for Computational Linguistics",
    url = "https://aclanthology.org/2024.acl-long.444/",
    doi = "10.18653/v1/2024.acl-long.444",
    pages = "8173--8186"
}

@inproceedings{xu-etal-2024-knowledge-conflicts,
    title = {{Knowledge Conflicts for {LLM}s: A Survey}},
    author = "Xu, Rongwu  and
      Qi, Zehan  and
      Guo, Zhijiang  and
      Wang, Cunxiang  and
      Wang, Hongru  and
      Zhang, Yue  and
      Xu, Wei",
    editor = "Al-Onaizan, Yaser  and
      Bansal, Mohit  and
      Chen, Yun-Nung",
    booktitle = "Proceedings of the 2024 Conference on Empirical Methods in Natural Language Processing",
    month = nov,
    year = "2024",
    address = "Miami, Florida, USA",
    publisher = "Association for Computational Linguistics",
    url = "https://aclanthology.org/2024.emnlp-main.486/",
    doi = "10.18653/v1/2024.emnlp-main.486",
    pages = "8541--8565"
}

@inproceedings{
wei2022chain,
title={{Chain of Thought Prompting Elicits Reasoning in Large Language Models}},
author={Jason Wei and Xuezhi Wang and Dale Schuurmans and Maarten Bosma and brian ichter and Fei Xia and Ed H. Chi and Quoc V Le and Denny Zhou},
booktitle={Advances in Neural Information Processing Systems},
editor={Alice H. Oh and Alekh Agarwal and Danielle Belgrave and Kyunghyun Cho},
year={2022},
url={https://openreview.net/forum?id=_VjQlMeSB_J}
}

@inproceedings{wang-atanasova-2025-self,
    title = {{Self-Critique and Refinement for Faithful Natural Language Explanations}},
    author = "Wang, Yingming  and
      Atanasova, Pepa",
    editor = "Christodoulopoulos, Christos  and
      Chakraborty, Tanmoy  and
      Rose, Carolyn  and
      Peng, Violet",
    booktitle = "Proceedings of the 2025 Conference on Empirical Methods in Natural Language Processing",
    month = nov,
    year = "2025",
    address = "Suzhou, China",
    publisher = "Association for Computational Linguistics",
    url = "https://aclanthology.org/2025.emnlp-main.427/",
    doi = "10.18653/v1/2025.emnlp-main.427",
    pages = "8481--8507",
    ISBN = "979-8-89176-332-6"
}

@inproceedings{wang-etal-2025-continuously,
    title = {{Continuously Steering {LLM}s Sensitivity to Contextual Knowledge with Proxy Models}},
    author = "Wang, Yilin  and
      Wang, Heng  and
      Bai, Yuyang  and
      Luo, Minnan",
    editor = "Christodoulopoulos, Christos  and
      Chakraborty, Tanmoy  and
      Rose, Carolyn  and
      Peng, Violet",
    booktitle = "Proceedings of the 2025 Conference on Empirical Methods in Natural Language Processing",
    month = nov,
    year = "2025",
    address = "Suzhou, China",
    publisher = "Association for Computational Linguistics",
    url = "https://aclanthology.org/2025.emnlp-main.233/",
    doi = "10.18653/v1/2025.emnlp-main.233",
    pages = "4682--4698",
    ISBN = "979-8-89176-332-6"
}

@inproceedings{wang-shu-2023-explainable,
    title = {{Explainable Claim Verification via Knowledge-Grounded Reasoning with Large Language Models}},
    author = "Wang, Haoran  and
      Shu, Kai",
    editor = "Bouamor, Houda  and
      Pino, Juan  and
      Bali, Kalika",
    booktitle = "Findings of the Association for Computational Linguistics: EMNLP 2023",
    month = dec,
    year = "2023",
    address = "Singapore",
    publisher = "Association for Computational Linguistics",
    url = "https://aclanthology.org/2023.findings-emnlp.416/",
    doi = "10.18653/v1/2023.findings-emnlp.416",
    pages = "6288--6304"
}

@Inbook{Wall2003,
author="Wall, Michael E.
and Rechtsteiner, Andreas
and Rocha, Luis M.",
editor="Berrar, Daniel P.
and Dubitzky, Werner
and Granzow, Martin",
title={{Singular Value Decomposition and Principal Component Analysis}},
bookTitle="A Practical Approach to Microarray Data Analysis",
year="2003",
publisher="Springer US",
address="Boston, MA",
pages="91--109",
isbn="978-0-306-47815-4",
doi="10.1007/0-306-47815-3_5",
url="https://doi.org/10.1007/0-306-47815-3_5"
}

@inproceedings{
turpin2023language,
title={{Language Models Don't Always Say What They Think: Unfaithful Explanations in Chain-of-Thought Prompting}},
author={Miles Turpin and Julian Michael and Ethan Perez and Samuel R. Bowman},
booktitle={Thirty-seventh Conference on Neural Information Processing Systems},
year={2023},
url={https://openreview.net/forum?id=bzs4uPLXvi}
}

@inproceedings{tenney-etal-2019-bert,
    title = {{{BERT} Rediscovers the Classical {NLP} Pipeline}},
    author = "Tenney, Ian  and
      Das, Dipanjan  and
      Pavlick, Ellie",
    editor = "Korhonen, Anna  and
      Traum, David  and
      M{\`a}rquez, Llu{\'i}s",
    booktitle = "Proceedings of the 57th Annual Meeting of the Association for Computational Linguistics",
    month = jul,
    year = "2019",
    address = "Florence, Italy",
    publisher = "Association for Computational Linguistics",
    url = "https://aclanthology.org/P19-1452/",
    doi = "10.18653/v1/P19-1452",
    pages = "4593--4601"
}

@misc{gemmateam2024gemma2improvingopen,
      title={{Gemma 2: Improving Open Language Models at a Practical Size}}, 
      author={Gemma Team and Morgane Riviere and Shreya Pathak and Pier Giuseppe Sessa and Cassidy Hardin and Surya Bhupatiraju and Léonard Hussenot and Thomas Mesnard and Bobak Shahriari and Alexandre Ramé and Johan Ferret and Peter Liu and Pouya Tafti and Abe Friesen and Michelle Casbon and Sabela Ramos and Ravin Kumar and Charline Le Lan and Sammy Jerome and Anton Tsitsulin and Nino Vieillard and Piotr Stanczyk and Sertan Girgin and Nikola Momchev and Matt Hoffman and Shantanu Thakoor and Jean-Bastien Grill and Behnam Neyshabur and Olivier Bachem and Alanna Walton and Aliaksei Severyn and Alicia Parrish and Aliya Ahmad and Allen Hutchison and Alvin Abdagic and Amanda Carl and Amy Shen and Andy Brock and Andy Coenen and Anthony Laforge and Antonia Paterson and Ben Bastian and Bilal Piot and Bo Wu and Brandon Royal and Charlie Chen and Chintu Kumar and Chris Perry and Chris Welty and Christopher A. Choquette-Choo and Danila Sinopalnikov and David Weinberger and Dimple Vijaykumar and Dominika Rogozińska and Dustin Herbison and Elisa Bandy and Emma Wang and Eric Noland and Erica Moreira and Evan Senter and Evgenii Eltyshev and Francesco Visin and Gabriel Rasskin and Gary Wei and Glenn Cameron and Gus Martins and Hadi Hashemi and Hanna Klimczak-Plucińska and Harleen Batra and Harsh Dhand and Ivan Nardini and Jacinda Mein and Jack Zhou and James Svensson and Jeff Stanway and Jetha Chan and Jin Peng Zhou and Joana Carrasqueira and Joana Iljazi and Jocelyn Becker and Joe Fernandez and Joost van Amersfoort and Josh Gordon and Josh Lipschultz and Josh Newlan and Ju-yeong Ji and Kareem Mohamed and Kartikeya Badola and Kat Black and Katie Millican and Keelin McDonell and Kelvin Nguyen and Kiranbir Sodhia and Kish Greene and Lars Lowe Sjoesund and Lauren Usui and Laurent Sifre and Lena Heuermann and Leticia Lago and Lilly McNealus and Livio Baldini Soares and Logan Kilpatrick and Lucas Dixon and Luciano Martins and Machel Reid and Manvinder Singh and Mark Iverson and Martin Görner and Mat Velloso and Mateo Wirth and Matt Davidow and Matt Miller and Matthew Rahtz and Matthew Watson and Meg Risdal and Mehran Kazemi and Michael Moynihan and Ming Zhang and Minsuk Kahng and Minwoo Park and Mofi Rahman and Mohit Khatwani and Natalie Dao and Nenshad Bardoliwalla and Nesh Devanathan and Neta Dumai and Nilay Chauhan and Oscar Wahltinez and Pankil Botarda and Parker Barnes and Paul Barham and Paul Michel and Pengchong Jin and Petko Georgiev and Phil Culliton and Pradeep Kuppala and Ramona Comanescu and Ramona Merhej and Reena Jana and Reza Ardeshir Rokni and Rishabh Agarwal and Ryan Mullins and Samaneh Saadat and Sara Mc Carthy and Sarah Cogan and Sarah Perrin and Sébastien M. R. Arnold and Sebastian Krause and Shengyang Dai and Shruti Garg and Shruti Sheth and Sue Ronstrom and Susan Chan and Timothy Jordan and Ting Yu and Tom Eccles and Tom Hennigan and Tomas Kocisky and Tulsee Doshi and Vihan Jain and Vikas Yadav and Vilobh Meshram and Vishal Dharmadhikari and Warren Barkley and Wei Wei and Wenming Ye and Woohyun Han and Woosuk Kwon and Xiang Xu and Zhe Shen and Zhitao Gong and Zichuan Wei and Victor Cotruta and Phoebe Kirk and Anand Rao and Minh Giang and Ludovic Peran and Tris Warkentin and Eli Collins and Joelle Barral and Zoubin Ghahramani and Raia Hadsell and D. Sculley and Jeanine Banks and Anca Dragan and Slav Petrov and Oriol Vinyals and Jeff Dean and Demis Hassabis and Koray Kavukcuoglu and Clement Farabet and Elena Buchatskaya and Sebastian Borgeaud and Noah Fiedel and Armand Joulin and Kathleen Kenealy and Robert Dadashi and Alek Andreev},
      year={2024},
      eprint={2408.00118},
      archivePrefix={arXiv},
      primaryClass={cs.CL},
      url={https://arxiv.org/abs/2408.00118}, 
}

@misc{tao2025lostinthelaterframeworkquantifyingcontextual,
      title={{"Lost-in-the-Later": Framework for Quantifying Contextual Grounding in Large Language Models}}, 
      author={Yufei Tao and Adam Hiatt and Rahul Seetharaman and Ameeta Agrawal},
      year={2025},
      eprint={2507.05424},
      archivePrefix={arXiv},
      primaryClass={cs.CL},
      url={https://arxiv.org/abs/2507.05424}, 
}

@inproceedings{tao-etal-2024-context,
    title = {{When Context Leads but Parametric Memory Follows in Large Language Models}},
    author = "Tao, Yufei  and
      Hiatt, Adam  and
      Haake, Erik  and
      Jetter, Antonie J.  and
      Agrawal, Ameeta",
    editor = "Al-Onaizan, Yaser  and
      Bansal, Mohit  and
      Chen, Yun-Nung",
    booktitle = "Proceedings of the 2024 Conference on Empirical Methods in Natural Language Processing",
    month = nov,
    year = "2024",
    address = "Miami, Florida, USA",
    publisher = "Association for Computational Linguistics",
    url = "https://aclanthology.org/2024.emnlp-main.234/",
    doi = "10.18653/v1/2024.emnlp-main.234",
    pages = "4034--4058"
}

@inproceedings{tan-etal-2025-improving,
    title = {{Improving Explainable Fact-Checking with Claim-Evidence Correlations}},
    author = "Tan, Xin  and
      Zou, Bowei  and
      Aw, Ai Ti",
    editor = "Rambow, Owen  and
      Wanner, Leo  and
      Apidianaki, Marianna  and
      Al-Khalifa, Hend  and
      Eugenio, Barbara Di  and
      Schockaert, Steven",
    booktitle = "Proceedings of the 31st International Conference on Computational Linguistics",
    month = jan,
    year = "2025",
    address = "Abu Dhabi, UAE",
    publisher = "Association for Computational Linguistics",
    url = "https://aclanthology.org/2025.coling-main.108/",
    pages = "1600--1612"
}

@inproceedings{
sun2025redeep,
title={{ReDe{EP}: Detecting Hallucination in Retrieval-Augmented Generation via Mechanistic Interpretability}},
author={ZhongXiang Sun and Xiaoxue Zang and Kai Zheng and Jun Xu and Xiao Zhang and Weijie Yu and Yang Song and Han Li},
booktitle={The Thirteenth International Conference on Learning Representations},
year={2025},
url={https://openreview.net/forum?id=ztzZDzgfrh}
}

@article{10.1162/COLI.a.621,
    author = {Sun, Jingyi and Atanasova, Pepa and Choudhury, Sagnik Ray and Islam, Sekh Mainul and Augenstein, Isabelle},
    title = {{Evaluation Framework for Highlight Explanations of Context Utilization in Language Models}},
    journal = {Computational Linguistics},
    pages = {1-35},
    year = {2026},
    month = {08},
    issn = {0891-2017},
    doi = {10.1162/COLI.a.621},
    url = {https://doi.org/10.1162/COLI.a.621},
    eprint = {https://direct.mit.edu/coli/article-pdf/doi/10.1162/COLI.a.621/2595375/coli.a.621.pdf},
}

@inproceedings{su-etal-2024-semi,
    title = {{Semi-Structured Chain-of-Thought: Integrating Multiple Sources of Knowledge for Improved Language Model Reasoning}},
    author = "Su, Xin  and
      Le, Tiep  and
      Bethard, Steven  and
      Howard, Phillip",
    editor = "Duh, Kevin  and
      Gomez, Helena  and
      Bethard, Steven",
    booktitle = "Proceedings of the 2024 Conference of the North American Chapter of the Association for Computational Linguistics: Human Language Technologies (Volume 1: Long Papers)",
    month = jun,
    year = "2024",
    address = "Mexico City, Mexico",
    publisher = "Association for Computational Linguistics",
    url = "https://aclanthology.org/2024.naacl-long.475/",
    doi = "10.18653/v1/2024.naacl-long.475",
    pages = "8597--8613"
}

@inproceedings{siegel-etal-2024-probabilities,
    title = {{The Probabilities Also Matter: A More Faithful Metric for Faithfulness of Free-Text Explanations in Large Language Models}},
    author = "Siegel, Noah  and
      Camburu, Oana-Maria  and
      Heess, Nicolas  and
      Perez-Ortiz, Maria",
    editor = "Ku, Lun-Wei  and
      Martins, Andre  and
      Srikumar, Vivek",
    booktitle = "Proceedings of the 62nd Annual Meeting of the Association for Computational Linguistics (Volume 2: Short Papers)",
    month = aug,
    year = "2024",
    address = "Bangkok, Thailand",
    publisher = "Association for Computational Linguistics",
    url = "https://aclanthology.org/2024.acl-short.49/",
    doi = "10.18653/v1/2024.acl-short.49",
    pages = "530--546"
}

@inproceedings{roberts-etal-2020-much,
    title = {{How Much Knowledge Can You Pack Into the Parameters of a Language Model?}},
    author = "Roberts, Adam  and
      Raffel, Colin  and
      Shazeer, Noam",
    editor = "Webber, Bonnie  and
      Cohn, Trevor  and
      He, Yulan  and
      Liu, Yang",
    booktitle = "Proceedings of the 2020 Conference on Empirical Methods in Natural Language Processing (EMNLP)",
    month = nov,
    year = "2020",
    address = "Online",
    publisher = "Association for Computational Linguistics",
    url = "https://aclanthology.org/2020.emnlp-main.437/",
    doi = "10.18653/v1/2020.emnlp-main.437",
    pages = "5418--5426"
}

@inproceedings{rajani-etal-2019-explain,
    title = {{Explain Yourself! Leveraging Language Models for Commonsense Reasoning}},
    author = "Rajani, Nazneen Fatema  and
      McCann, Bryan  and
      Xiong, Caiming  and
      Socher, Richard",
    editor = "Korhonen, Anna  and
      Traum, David  and
      M{\`a}rquez, Llu{\'i}s",
    booktitle = "Proceedings of the 57th Annual Meeting of the Association for Computational Linguistics",
    month = jul,
    year = "2019",
    address = "Florence, Italy",
    publisher = "Association for Computational Linguistics",
    url = "https://aclanthology.org/P19-1487/",
    doi = "10.18653/v1/P19-1487",
    pages = "4932--4942"
}

@inproceedings{petroni-etal-2019-language,
    title = {{Language Models as Knowledge Bases?}},
    author = {Petroni, Fabio  and
      Rockt{\"a}schel, Tim  and
      Riedel, Sebastian  and
      Lewis, Patrick  and
      Bakhtin, Anton  and
      Wu, Yuxiang  and
      Miller, Alexander},
    editor = "Inui, Kentaro  and
      Jiang, Jing  and
      Ng, Vincent  and
      Wan, Xiaojun",
    booktitle = "Proceedings of the 2019 Conference on Empirical Methods in Natural Language Processing and the 9th International Joint Conference on Natural Language Processing (EMNLP-IJCNLP)",
    month = nov,
    year = "2019",
    address = "Hong Kong, China",
    publisher = "Association for Computational Linguistics",
    url = "https://aclanthology.org/D19-1250/",
    doi = "10.18653/v1/D19-1250",
    pages = "2463--2473"
}

@inproceedings{
minder2025controllable,
title={{Controllable Context Sensitivity and the Knob Behind It}},
author={Julian Minder and Kevin Du and Niklas Stoehr and Giovanni Monea and Chris Wendler and Robert West and Ryan Cotterell},
booktitle={The Thirteenth International Conference on Learning Representations},
year={2025},
url={https://openreview.net/forum?id=Igm9bbkzHC}
}

@inproceedings{mihaylov-etal-2018-suit,
    title = {{Can a Suit of Armor Conduct Electricity? A New Dataset for Open Book Question Answering}},
    author = "Mihaylov, Todor  and
      Clark, Peter  and
      Khot, Tushar  and
      Sabharwal, Ashish",
    editor = "Riloff, Ellen  and
      Chiang, David  and
      Hockenmaier, Julia  and
      Tsujii, Jun{'}ichi",
    booktitle = "Proceedings of the 2018 Conference on Empirical Methods in Natural Language Processing",
    month = oct # "-" # nov,
    year = "2018",
    address = "Brussels, Belgium",
    publisher = "Association for Computational Linguistics",
    url = "https://aclanthology.org/D18-1260/",
    doi = "10.18653/v1/D18-1260",
    pages = "2381--2391"
}

@inproceedings{marjanovic-etal-2024-dynamicqa,
    title = {{{DYNAMICQA}: Tracing Internal Knowledge Conflicts in Language Models}},
    author = "Marjanovic, Sara Vera  and
      Yu, Haeun  and
      Atanasova, Pepa  and
      Maistro, Maria  and
      Lioma, Christina  and
      Augenstein, Isabelle",
    editor = "Al-Onaizan, Yaser  and
      Bansal, Mohit  and
      Chen, Yun-Nung",
    booktitle = "Findings of the Association for Computational Linguistics: EMNLP 2024",
    month = nov,
    year = "2024",
    address = "Miami, Florida, USA",
    publisher = "Association for Computational Linguistics",
    url = "https://aclanthology.org/2024.findings-emnlp.838/",
    doi = "10.18653/v1/2024.findings-emnlp.838",
    pages = "14346--14360"
}

@inproceedings{longpre-etal-2021-entity,
    title = {{Entity-Based Knowledge Conflicts in Question Answering}},
    author = "Longpre, Shayne  and
      Perisetla, Kartik  and
      Chen, Anthony  and
      Ramesh, Nikhil  and
      DuBois, Chris  and
      Singh, Sameer",
    editor = "Moens, Marie-Francine  and
      Huang, Xuanjing  and
      Specia, Lucia  and
      Yih, Scott Wen-tau",
    booktitle = "Proceedings of the 2021 Conference on Empirical Methods in Natural Language Processing",
    month = nov,
    year = "2021",
    address = "Online and Punta Cana, Dominican Republic",
    publisher = "Association for Computational Linguistics",
    url = "https://aclanthology.org/2021.emnlp-main.565/",
    doi = "10.18653/v1/2021.emnlp-main.565",
    pages = "7052--7063"
}

@article{JMLR:v26:24-0699,
  author  = {Margherita Lazzaretto and Jonas Peters and Niklas Pfister},
  title   = {{Invariant Subspace Decomposition}},
  journal = {Journal of Machine Learning Research},
  year    = {2025},
  volume  = {26},
  number  = {95},
  pages   = {1--56},
  url     = {http://jmlr.org/papers/v26/24-0699.html}
}

@inproceedings{lampinen-etal-2022-language,
    title = {{Can language models learn from explanations in context?}},
    author = "Lampinen, Andrew  and
      Dasgupta, Ishita  and
      Chan, Stephanie  and
      Mathewson, Kory  and
      Tessler, Mh  and
      Creswell, Antonia  and
      McClelland, James  and
      Wang, Jane  and
      Hill, Felix",
    editor = "Goldberg, Yoav  and
      Kozareva, Zornitsa  and
      Zhang, Yue",
    booktitle = "Findings of the Association for Computational Linguistics: EMNLP 2022",
    month = dec,
    year = "2022",
    address = "Abu Dhabi, United Arab Emirates",
    publisher = "Association for Computational Linguistics",
    url = "https://aclanthology.org/2022.findings-emnlp.38/",
    doi = "10.18653/v1/2022.findings-emnlp.38",
    pages = "537--563"
}

@article{jiang-etal-2020-know,
    title = {{How Can We Know What Language Models Know?}},
    author = "Jiang, Zhengbao  and
      Xu, Frank F.  and
      Araki, Jun  and
      Neubig, Graham",
    editor = "Johnson, Mark  and
      Roark, Brian  and
      Nenkova, Ani",
    journal = "Transactions of the Association for Computational Linguistics",
    volume = "8",
    year = "2020",
    address = "Cambridge, MA",
    publisher = "MIT Press",
    url = "https://aclanthology.org/2020.tacl-1.28/",
    doi = "10.1162/tacl_a_00324",
    pages = "423--438"
}

@misc{jiang2023mistral7b,
      title={{Mistral 7B}}, 
      author={Albert Q. Jiang and Alexandre Sablayrolles and Arthur Mensch and Chris Bamford and Devendra Singh Chaplot and Diego de las Casas and Florian Bressand and Gianna Lengyel and Guillaume Lample and Lucile Saulnier and Lélio Renard Lavaud and Marie-Anne Lachaux and Pierre Stock and Teven Le Scao and Thibaut Lavril and Thomas Wang and Timothée Lacroix and William El Sayed},
      year={2023},
      eprint={2310.06825},
      archivePrefix={arXiv},
      primaryClass={cs.CL},
      url={https://arxiv.org/abs/2310.06825}, 
}

@inproceedings{hu-etal-2024-knowledge,
    title = {{Knowledge-Centric Hallucination Detection}},
    author = "Hu, Xiangkun  and
      Ru, Dongyu  and
      Qiu, Lin  and
      Guo, Qipeng  and
      Zhang, Tianhang  and
      Xu, Yang  and
      Luo, Yun  and
      Liu, Pengfei  and
      Zhang, Yue  and
      Zhang, Zheng",
    editor = "Al-Onaizan, Yaser  and
      Bansal, Mohit  and
      Chen, Yun-Nung",
    booktitle = "Proceedings of the 2024 Conference on Empirical Methods in Natural Language Processing",
    month = nov,
    year = "2024",
    address = "Miami, Florida, USA",
    publisher = "Association for Computational Linguistics",
    url = "https://aclanthology.org/2024.emnlp-main.395/",
    doi = "10.18653/v1/2024.emnlp-main.395",
    pages = "6953--6975"
}

@inproceedings{hewitt-manning-2019-structural,
    title = {{{A} Structural Probe for Finding Syntax in Word Representations}},
    author = "Hewitt, John  and
      Manning, Christopher D.",
    editor = "Burstein, Jill  and
      Doran, Christy  and
      Solorio, Thamar",
    booktitle = "Proceedings of the 2019 Conference of the North {A}merican Chapter of the Association for Computational Linguistics: Human Language Technologies, Volume 1 (Long and Short Papers)",
    month = jun,
    year = "2019",
    address = "Minneapolis, Minnesota",
    publisher = "Association for Computational Linguistics",
    url = "https://aclanthology.org/N19-1419/",
    doi = "10.18653/v1/N19-1419",
    pages = "4129--4138"
}

@inproceedings{hagstrom-etal-2025-reality,
    title = {{A Reality Check on Context Utilisation for Retrieval-Augmented Generation}},
    author = {Hagstr{\"o}m, Lovisa  and
      Marjanovic, Sara Vera  and
      Yu, Haeun  and
      Arora, Arnav  and
      Lioma, Christina  and
      Maistro, Maria  and
      Atanasova, Pepa  and
      Augenstein, Isabelle},
    editor = "Che, Wanxiang  and
      Nabende, Joyce  and
      Shutova, Ekaterina  and
      Pilehvar, Mohammad Taher",
    booktitle = "Proceedings of the 63rd Annual Meeting of the Association for Computational Linguistics (Volume 1: Long Papers)",
    month = jul,
    year = "2025",
    address = "Vienna, Austria",
    publisher = "Association for Computational Linguistics",
    url = "https://aclanthology.org/2025.acl-long.968/",
    doi = "10.18653/v1/2025.acl-long.968",
    pages = "19691--19730",
    ISBN = "979-8-89176-251-0"
}

@inproceedings{hagstrom-etal-2026-cub,
    title = {{{CUB}: Benchmarking Context Utilisation Techniques for Language Models}},
    author = {Hagstr{\"o}m, Lovisa  and
      Kim, Youna  and
      Yu, Haeun  and
      Lee, Sang-goo  and
      Johansson, Richard  and
      Cho, Hyunsoo  and
      Augenstein, Isabelle},
    editor = "Liakata, Maria  and
      Moreira, Viviane P.  and
      Zhang, Jiajun  and
      Jurgens, David",
    booktitle = "Proceedings of the 64th Annual Meeting of the {A}ssociation for {C}omputational {L}inguistics (Volume 1: Long Papers)",
    month = jul,
    year = "2026",
    address = "San Diego, California, United States",
    publisher = "Association for Computational Linguistics",
    url = "https://aclanthology.org/2026.acl-long.1151/",
    doi = "10.18653/v1/2026.acl-long.1151",
    pages = "25101--25133",
    ISBN = "979-8-89176-390-6"
}

@misc{grattafiori2024llama3herdmodels,
      title={{The Llama 3 Herd of Models}}, 
      author={Aaron Grattafiori and Abhimanyu Dubey and Abhinav Jauhri and Abhinav Pandey and Abhishek Kadian and Ahmad Al-Dahle and Aiesha Letman and Akhil Mathur and Alan Schelten and Alex Vaughan and Amy Yang and Angela Fan and Anirudh Goyal and Anthony Hartshorn and Aobo Yang and Archi Mitra and Archie Sravankumar and Artem Korenev and Arthur Hinsvark and Arun Rao and Aston Zhang and Aurelien Rodriguez and Austen Gregerson and Ava Spataru and Baptiste Roziere and Bethany Biron and Binh Tang and Bobbie Chern and Charlotte Caucheteux and Chaya Nayak and Chloe Bi and Chris Marra and Chris McConnell and Christian Keller and Christophe Touret and Chunyang Wu and Corinne Wong and Cristian Canton Ferrer and Cyrus Nikolaidis and Damien Allonsius and Daniel Song and Danielle Pintz and Danny Livshits and Danny Wyatt and David Esiobu and Dhruv Choudhary and Dhruv Mahajan and Diego Garcia-Olano and Diego Perino and Dieuwke Hupkes and Egor Lakomkin and Ehab AlBadawy and Elina Lobanova and Emily Dinan and Eric Michael Smith and Filip Radenovic and Francisco Guzmán and Frank Zhang and Gabriel Synnaeve and Gabrielle Lee and Georgia Lewis Anderson and Govind Thattai and Graeme Nail and Gregoire Mialon and Guan Pang and Guillem Cucurell and Hailey Nguyen and Hannah Korevaar and Hu Xu and Hugo Touvron and Iliyan Zarov and Imanol Arrieta Ibarra and Isabel Kloumann and Ishan Misra and Ivan Evtimov and Jack Zhang and Jade Copet and Jaewon Lee and Jan Geffert and Jana Vranes and Jason Park and Jay Mahadeokar and Jeet Shah and Jelmer van der Linde and Jennifer Billock and Jenny Hong and Jenya Lee and Jeremy Fu and Jianfeng Chi and Jianyu Huang and Jiawen Liu and Jie Wang and Jiecao Yu and Joanna Bitton and Joe Spisak and Jongsoo Park and Joseph Rocca and Joshua Johnstun and Joshua Saxe and Junteng Jia and Kalyan Vasuden Alwala and Karthik Prasad and Kartikeya Upasani and Kate Plawiak and Ke Li and Kenneth Heafield and Kevin Stone and Khalid El-Arini and Krithika Iyer and Kshitiz Malik and Kuenley Chiu and Kunal Bhalla and Kushal Lakhotia and Lauren Rantala-Yeary and Laurens van der Maaten and Lawrence Chen and Liang Tan and Liz Jenkins and Louis Martin and Lovish Madaan and Lubo Malo and Lukas Blecher and Lukas Landzaat and Luke de Oliveira and Madeline Muzzi and Mahesh Pasupuleti and Mannat Singh and Manohar Paluri and Marcin Kardas and Maria Tsimpoukelli and Mathew Oldham and Mathieu Rita and Maya Pavlova and Melanie Kambadur and Mike Lewis and Min Si and Mitesh Kumar Singh and Mona Hassan and Naman Goyal and Narjes Torabi and Nikolay Bashlykov and Nikolay Bogoychev and Niladri Chatterji and Ning Zhang and Olivier Duchenne and Onur Çelebi and Patrick Alrassy and Pengchuan Zhang and Pengwei Li and Petar Vasic and Peter Weng and Prajjwal Bhargava and Pratik Dubal and Praveen Krishnan and Punit Singh Koura and Puxin Xu and Qing He and Qingxiao Dong and Ragavan Srinivasan and Raj Ganapathy and Ramon Calderer and Ricardo Silveira Cabral and Robert Stojnic and Roberta Raileanu and Rohan Maheswari and Rohit Girdhar and Rohit Patel and Romain Sauvestre and Ronnie Polidoro and Roshan Sumbaly and Ross Taylor and Ruan Silva and Rui Hou and Rui Wang and Saghar Hosseini and Sahana Chennabasappa and Sanjay Singh and Sean Bell and Seohyun Sonia Kim and Sergey Edunov and Shaoliang Nie and Sharan Narang and Sharath Raparthy and Sheng Shen and Shengye Wan and Shruti Bhosale and Shun Zhang and Simon Vandenhende and Soumya Batra and Spencer Whitman and Sten Sootla and Stephane Collot and Suchin Gururangan and Sydney Borodinsky and Tamar Herman and Tara Fowler and Tarek Sheasha and Thomas Georgiou and Thomas Scialom and Tobias Speckbacher and Todor Mihaylov and Tong Xiao and Ujjwal Karn and Vedanuj Goswami and Vibhor Gupta and Vignesh Ramanathan and Viktor Kerkez and Vincent Gonguet and Virginie Do and Vish Vogeti and Vítor Albiero and Vladan Petrovic and Weiwei Chu and Wenhan Xiong and Wenyin Fu and Whitney Meers and Xavier Martinet and Xiaodong Wang and Xiaofang Wang and Xiaoqing Ellen Tan and Xide Xia and Xinfeng Xie and Xuchao Jia and Xuewei Wang and Yaelle Goldschlag and Yashesh Gaur and Yasmine Babaei and Yi Wen and Yiwen Song and Yuchen Zhang and Yue Li and Yuning Mao and Zacharie Delpierre Coudert and Zheng Yan and Zhengxing Chen and Zoe Papakipos and Aaditya Singh and Aayushi Srivastava and Abha Jain and Adam Kelsey and Adam Shajnfeld and Adithya Gangidi and Adolfo Victoria and Ahuva Goldstand and Ajay Menon and Ajay Sharma and Alex Boesenberg and Alexei Baevski and Allie Feinstein and Amanda Kallet and Amit Sangani and Amos Teo and Anam Yunus and Andrei Lupu and Andres Alvarado and Andrew Caples and Andrew Gu and Andrew Ho and Andrew Poulton and Andrew Ryan and Ankit Ramchandani and Annie Dong and Annie Franco and Anuj Goyal and Aparajita Saraf and Arkabandhu Chowdhury and Ashley Gabriel and Ashwin Bharambe and Assaf Eisenman and Azadeh Yazdan and Beau James and Ben Maurer and Benjamin Leonhardi and Bernie Huang and Beth Loyd and Beto De Paola and Bhargavi Paranjape and Bing Liu and Bo Wu and Boyu Ni and Braden Hancock and Bram Wasti and Brandon Spence and Brani Stojkovic and Brian Gamido and Britt Montalvo and Carl Parker and Carly Burton and Catalina Mejia and Ce Liu and Changhan Wang and Changkyu Kim and Chao Zhou and Chester Hu and Ching-Hsiang Chu and Chris Cai and Chris Tindal and Christoph Feichtenhofer and Cynthia Gao and Damon Civin and Dana Beaty and Daniel Kreymer and Daniel Li and David Adkins and David Xu and Davide Testuggine and Delia David and Devi Parikh and Diana Liskovich and Didem Foss and Dingkang Wang and Duc Le and Dustin Holland and Edward Dowling and Eissa Jamil and Elaine Montgomery and Eleonora Presani and Emily Hahn and Emily Wood and Eric-Tuan Le and Erik Brinkman and Esteban Arcaute and Evan Dunbar and Evan Smothers and Fei Sun and Felix Kreuk and Feng Tian and Filippos Kokkinos and Firat Ozgenel and Francesco Caggioni and Frank Kanayet and Frank Seide and Gabriela Medina Florez and Gabriella Schwarz and Gada Badeer and Georgia Swee and Gil Halpern and Grant Herman and Grigory Sizov and Guangyi and Zhang and Guna Lakshminarayanan and Hakan Inan and Hamid Shojanazeri and Han Zou and Hannah Wang and Hanwen Zha and Haroun Habeeb and Harrison Rudolph and Helen Suk and Henry Aspegren and Hunter Goldman and Hongyuan Zhan and Ibrahim Damlaj and Igor Molybog and Igor Tufanov and Ilias Leontiadis and Irina-Elena Veliche and Itai Gat and Jake Weissman and James Geboski and James Kohli and Janice Lam and Japhet Asher and Jean-Baptiste Gaya and Jeff Marcus and Jeff Tang and Jennifer Chan and Jenny Zhen and Jeremy Reizenstein and Jeremy Teboul and Jessica Zhong and Jian Jin and Jingyi Yang and Joe Cummings and Jon Carvill and Jon Shepard and Jonathan McPhie and Jonathan Torres and Josh Ginsburg and Junjie Wang and Kai Wu and Kam Hou U and Karan Saxena and Kartikay Khandelwal and Katayoun Zand and Kathy Matosich and Kaushik Veeraraghavan and Kelly Michelena and Keqian Li and Kiran Jagadeesh and Kun Huang and Kunal Chawla and Kyle Huang and Lailin Chen and Lakshya Garg and Lavender A and Leandro Silva and Lee Bell and Lei Zhang and Liangpeng Guo and Licheng Yu and Liron Moshkovich and Luca Wehrstedt and Madian Khabsa and Manav Avalani and Manish Bhatt and Martynas Mankus and Matan Hasson and Matthew Lennie and Matthias Reso and Maxim Groshev and Maxim Naumov and Maya Lathi and Meghan Keneally and Miao Liu and Michael L. Seltzer and Michal Valko and Michelle Restrepo and Mihir Patel and Mik Vyatskov and Mikayel Samvelyan and Mike Clark and Mike Macey and Mike Wang and Miquel Jubert Hermoso and Mo Metanat and Mohammad Rastegari and Munish Bansal and Nandhini Santhanam and Natascha Parks and Natasha White and Navyata Bawa and Nayan Singhal and Nick Egebo and Nicolas Usunier and Nikhil Mehta and Nikolay Pavlovich Laptev and Ning Dong and Norman Cheng and Oleg Chernoguz and Olivia Hart and Omkar Salpekar and Ozlem Kalinli and Parkin Kent and Parth Parekh and Paul Saab and Pavan Balaji and Pedro Rittner and Philip Bontrager and Pierre Roux and Piotr Dollar and Polina Zvyagina and Prashant Ratanchandani and Pritish Yuvraj and Qian Liang and Rachad Alao and Rachel Rodriguez and Rafi Ayub and Raghotham Murthy and Raghu Nayani and Rahul Mitra and Rangaprabhu Parthasarathy and Raymond Li and Rebekkah Hogan and Robin Battey and Rocky Wang and Russ Howes and Ruty Rinott and Sachin Mehta and Sachin Siby and Sai Jayesh Bondu and Samyak Datta and Sara Chugh and Sara Hunt and Sargun Dhillon and Sasha Sidorov and Satadru Pan and Saurabh Mahajan and Saurabh Verma and Seiji Yamamoto and Sharadh Ramaswamy and Shaun Lindsay and Shaun Lindsay and Sheng Feng and Shenghao Lin and Shengxin Cindy Zha and Shishir Patil and Shiva Shankar and Shuqiang Zhang and Shuqiang Zhang and Sinong Wang and Sneha Agarwal and Soji Sajuyigbe and Soumith Chintala and Stephanie Max and Stephen Chen and Steve Kehoe and Steve Satterfield and Sudarshan Govindaprasad and Sumit Gupta and Summer Deng and Sungmin Cho and Sunny Virk and Suraj Subramanian and Sy Choudhury and Sydney Goldman and Tal Remez and Tamar Glaser and Tamara Best and Thilo Koehler and Thomas Robinson and Tianhe Li and Tianjun Zhang and Tim Matthews and Timothy Chou and Tzook Shaked and Varun Vontimitta and Victoria Ajayi and Victoria Montanez and Vijai Mohan and Vinay Satish Kumar and Vishal Mangla and Vlad Ionescu and Vlad Poenaru and Vlad Tiberiu Mihailescu and Vladimir Ivanov and Wei Li and Wenchen Wang and Wenwen Jiang and Wes Bouaziz and Will Constable and Xiaocheng Tang and Xiaojian Wu and Xiaolan Wang and Xilun Wu and Xinbo Gao and Yaniv Kleinman and Yanjun Chen and Ye Hu and Ye Jia and Ye Qi and Yenda Li and Yilin Zhang and Ying Zhang and Yossi Adi and Youngjin Nam and Yu and Wang and Yu Zhao and Yuchen Hao and Yundi Qian and Yunlu Li and Yuzi He and Zach Rait and Zachary DeVito and Zef Rosnbrick and Zhaoduo Wen and Zhenyu Yang and Zhiwei Zhao and Zhiyu Ma},
      year={2024},
      eprint={2407.21783},
      archivePrefix={arXiv},
      primaryClass={cs.AI},
      url={https://arxiv.org/abs/2407.21783}, 
}

@article{geva-etal-2021-aristotle,
    title = {{Did Aristotle Use a Laptop? A Question Answering Benchmark with Implicit Reasoning Strategies}},
    author = "Geva, Mor  and
      Khashabi, Daniel  and
      Segal, Elad  and
      Khot, Tushar  and
      Roth, Dan  and
      Berant, Jonathan",
    editor = "Roark, Brian  and
      Nenkova, Ani",
    journal = "Transactions of the Association for Computational Linguistics",
    volume = "9",
    year = "2021",
    address = "Cambridge, MA",
    publisher = "MIT Press",
    url = "https://aclanthology.org/2021.tacl-1.21/",
    doi = "10.1162/tacl_a_00370",
    pages = "346--361"
}

@article{elhage2022superposition,
   title={Toy Models of Superposition},
   author={Elhage, Nelson and Hume, Tristan and Olsson, Catherine and Schiefer, Nicholas and Henighan, Tom and Kravec, Shauna and Hatfield-Dodds, Zac and Lasenby, Robert and Drain, Dawn and Chen, Carol and Grosse, Roger and McCandlish, Sam and Kaplan, Jared and Amodei, Dario and Wattenberg, Martin and Olah, Christopher},
   year={2022},
   journal={Transformer Circuits Thread},
   url={https://transformer-circuits.pub/2022/toy_model/index.html}
}

@inproceedings{clark-etal-2019-bert,
    title = {{What Does {BERT} Look at? An Analysis of {BERT}{'}s Attention}},
    author = "Clark, Kevin  and
      Khandelwal, Urvashi  and
      Levy, Omer  and
      Manning, Christopher D.",
    editor = "Linzen, Tal  and
      Chrupa{\l}a, Grzegorz  and
      Belinkov, Yonatan  and
      Hupkes, Dieuwke",
    booktitle = "Proceedings of the 2019 ACL Workshop BlackboxNLP: Analyzing and Interpreting Neural Networks for NLP",
    month = aug,
    year = "2019",
    address = "Florence, Italy",
    publisher = "Association for Computational Linguistics",
    url = "https://aclanthology.org/W19-4828/",
    doi = "10.18653/v1/W19-4828",
    pages = "276--286"
}

@misc{cheng2024understandinginterplayparametriccontextual,
      title={{Understanding the Interplay between Parametric and Contextual Knowledge for Large Language Models}}, 
      author={Sitao Cheng and Liangming Pan and Xunjian Yin and Xinyi Wang and William Yang Wang},
      year={2024},
      eprint={2410.08414},
      archivePrefix={arXiv},
      primaryClass={cs.CL},
      url={https://arxiv.org/abs/2410.08414}, 
}

@inproceedings{NEURIPS2018_4c7a167b,
 author = {Camburu, Oana-Maria and Rockt\"{a}schel, Tim and Lukasiewicz, Thomas and Blunsom, Phil},
 booktitle = {Advances in Neural Information Processing Systems},
 editor = {S. Bengio and H. Wallach and H. Larochelle and K. Grauman and N. Cesa-Bianchi and R. Garnett},
 pages = {},
 publisher = {Curran Associates, Inc.},
 title = {{e-SNLI: Natural Language Inference with Natural Language Explanations}},
 url = {https://proceedings.neurips.cc/paper_files/paper/2018/file/4c7a167bb329bd92580a99ce422d6fa6-Paper.pdf},
 volume = {31},
 year = {2018}
}

@inproceedings{NEURIPS2020_1457c0d6,
 author = {Brown, Tom and Mann, Benjamin and Ryder, Nick and Subbiah, Melanie and Kaplan, Jared D and Dhariwal, Prafulla and Neelakantan, Arvind and Shyam, Pranav and Sastry, Girish and Askell, Amanda and Agarwal, Sandhini and Herbert-Voss, Ariel and Krueger, Gretchen and Henighan, Tom and Child, Rewon and Ramesh, Aditya and Ziegler, Daniel and Wu, Jeffrey and Winter, Clemens and Hesse, Chris and Chen, Mark and Sigler, Eric and Litwin, Mateusz and Gray, Scott and Chess, Benjamin and Clark, Jack and Berner, Christopher and McCandlish, Sam and Radford, Alec and Sutskever, Ilya and Amodei, Dario},
 booktitle = {Advances in Neural Information Processing Systems},
 editor = {H. Larochelle and M. Ranzato and R. Hadsell and M.F. Balcan and H. Lin},
 pages = {1877--1901},
 publisher = {Curran Associates, Inc.},
 title = {{Language Models are Few-Shot Learners}},
 url = {https://proceedings.neurips.cc/paper_files/paper/2020/file/1457c0d6bfcb4967418bfb8ac142f64a-Paper.pdf},
 volume = {33},
 year = {2020}
}

@inproceedings{
bi2026parameters,
title={{Parameters vs. Context: Fine-Grained Control of Knowledge Reliance in Language Models}},
author={Baolong Bi and Shenghua Liu and Yiwei Wang and Yilong Xu and Junfeng Fang and Lingrui Mei and Xueqi Cheng},
booktitle={The Fourteenth International Conference on Learning Representations},
year={2026},
url={https://openreview.net/forum?id=TJ3DqFiGau}
}

@inproceedings{atanasova-etal-2020-generating-fact,
    title = {{Generating Fact Checking Explanations}},
    author = "Atanasova, Pepa  and
      Simonsen, Jakob Grue  and
      Lioma, Christina  and
      Augenstein, Isabelle",
    editor = "Jurafsky, Dan  and
      Chai, Joyce  and
      Schluter, Natalie  and
      Tetreault, Joel",
    booktitle = "Proceedings of the 58th Annual Meeting of the Association for Computational Linguistics",
    month = jul,
    year = "2020",
    address = "Online",
    publisher = "Association for Computational Linguistics",
    url = "https://aclanthology.org/2020.acl-main.656/",
    doi = "10.18653/v1/2020.acl-main.656",
    pages = "7352--7364"
}

@inproceedings{atanasova-etal-2023-faithfulness,
    title = {{Faithfulness Tests for Natural Language Explanations}},
    author = "Atanasova, Pepa  and
      Camburu, Oana-Maria  and
      Lioma, Christina  and
      Lukasiewicz, Thomas  and
      Simonsen, Jakob Grue  and
      Augenstein, Isabelle",
    editor = "Rogers, Anna  and
      Boyd-Graber, Jordan  and
      Okazaki, Naoaki",
    booktitle = "Proceedings of the 61st Annual Meeting of the Association for Computational Linguistics (Volume 2: Short Papers)",
    month = jul,
    year = "2023",
    address = "Toronto, Canada",
    publisher = "Association for Computational Linguistics",
    url = "https://aclanthology.org/2023.acl-short.25/",
    doi = "10.18653/v1/2023.acl-short.25",
    pages = "283--294"
}

@inproceedings{niu-etal-2024-ragtruth,
    title = {{{RAGT}ruth: A Hallucination Corpus for Developing Trustworthy Retrieval-Augmented Language Models}},
    author = "Niu, Cheng  and
      Wu, Yuanhao  and
      Zhu, Juno  and
      Xu, Siliang  and
      Shum, KaShun  and
      Zhong, Randy  and
      Song, Juntong  and
      Zhang, Tong",
    editor = "Ku, Lun-Wei  and
      Martins, Andre  and
      Srikumar, Vivek",
    booktitle = "Proceedings of the 62nd Annual Meeting of the Association for Computational Linguistics (Volume 1: Long Papers)",
    month = aug,
    year = "2024",
    address = "Bangkok, Thailand",
    publisher = "Association for Computational Linguistics",
    url = "https://aclanthology.org/2024.acl-long.585/",
    doi = "10.18653/v1/2024.acl-long.585",
    pages = "10862--10878"
}

@InProceedings{pmlr-v235-ghandeharioun24a,
  title = 	 {{Patchscopes: A Unifying Framework for Inspecting Hidden Representations of Language Models}},
  author =       {Ghandeharioun, Asma and Caciularu, Avi and Pearce, Adam and Dixon, Lucas and Geva, Mor},
  booktitle = 	 {Proceedings of the 41st International Conference on Machine Learning},
  pages = 	 {15466--15490},
  year = 	 {2024},
  editor = 	 {Salakhutdinov, Ruslan and Kolter, Zico and Heller, Katherine and Weller, Adrian and Oliver, Nuria and Scarlett, Jonathan and Berkenkamp, Felix},
  volume = 	 {235},
  series = 	 {Proceedings of Machine Learning Research},
  month = 	 {21--27 Jul},
  publisher =    {PMLR},
  url = 	 {https://proceedings.mlr.press/v235/ghandeharioun24a.html}
}

@misc{wangsajaya2026emergencecontextcharacteristicssensitivity,
      title={{Emergence of Context Characteristics Sensitivity in Large Language Models}}, 
      author={Nadya Yuki Wangsajaya and Haeun Yu and Isabelle Augenstein},
      year={2026},
      eprint={2606.09525},
      archivePrefix={arXiv},
      primaryClass={cs.CL},
      url={https://arxiv.org/abs/2606.09525}, 
}

@misc{sun2026investigatinginterplaycontextualparametric,
      title={{Investigating the Interplay between Contextual and Parametric Chain-of-Thought Faithfulness under Optimization}}, 
      author={Jingyi Sun and Qianli Wang and Pepa Atanasova and Nils Feldhus and Isabelle Augenstein},
      year={2026},
      eprint={2605.24960},
      archivePrefix={arXiv},
      primaryClass={cs.CL},
      url={https://arxiv.org/abs/2605.24960}, 
}

\clearpage
\appendix

\section{Appendix}
\label{sec:appendix}

\subsection{Replication Details}\label{app:reproducibility}



\subsubsection{Determining PK and CK Directions in the Rank-2 Projection Subspace.}\label{app:pk_ck_direction_assignment} Once we obtain the rank-2 projection subspace $\mathbf{P}$ spanned by the orthonormal basis vectors $\vec{\mathbf{u}} = [\vec{\mathbf{u}}_c,\vec{\mathbf{u}}_p] \in \mathbb{R}^{d \times 2}$, we identify the PK and CK directions as follows:
\begin{align}
    \vec{\mathbf{u}}_{\mathrm{p}} &= \arg\max_{\vec{\mathbf{u}}_p, \vec{\mathbf{u}}_c} \left( \vec{\mathbf{u}}^\top \mathbf{H}_{w_p} \right), \label{eq:pk_dir} \\
    \vec{\mathbf{u}}_{\mathrm{c}} &= \arg\max_{\vec{\mathbf{u}}_p, \vec{\mathbf{u}}_c} \left( \vec{\mathbf{u}}^\top \mathbf{H}_{w_c} \right), \label{eq:ck_dir}
\end{align}

where $\mathbf{H}_{w_p}\in \mathbb{R}^{n\times d}$, and $\mathbf{H}_{w_c} \in \mathbb{R}^{n\times d}$ are matrices of hidden representation of answer tokens $a(q, \phi)$ and $a(q, c)$ guided by intents $w_p$ and $w_c$ respectively over $n$ examples. If the optimization for learning $\vec{\mathbf{u}}_p$ and $\vec{\mathbf{u}}_c$ (\S\ref{sec:u_learning_optimization}) converges well based on two pre-selected directions, then we will always get two different directions for the two knowledge sources.

\subsubsection{Prompt Template}\label{app:prompt_template}
Prompt template to generate controlled answers $a(q, \phi)$, $a(q, c)$ and $a$ driven by intents $w_p$, $w_c$, and $w_b$ are described in \cref{tab:app_intent_prompt_template}. The prompt template for generating NLEs and CoT-based prompts across datasets and models is described in \cref{tab:nle_cot_prompt_template}. 

\begin{table*}[!ht]
    \centering
    \begin{tabular}{l|p{13cm}}
        \toprule
        \textbf{Intent} & \textbf{Prompt} \\ \midrule
        $w_p$ &
        \texttt{<|begin\_of\_text|><|start\_header\_id|>system<|end\_header\_id|>}\newline
        Answer the following query considering the provided context. Answer with only one word.\texttt{<|eot\_id|><|start\_header\_id|>user<|end\_header\_id|>}\newline
        Context: Daily Grind Cafe is a popular place to eat lunch out. At a Potluck Party, everyone brings their own lunch to share with friends.\newline
        Instruction: Ignore the context in answering the query.\newline
        Query: Which action can help a person save money?\newline
        A. stop eating at Daily Grind Cafe\newline
        B. attend a Potluck Party\newline
        C. Unknown\texttt{<|eot\_id|><|start\_header\_id|>assistant<|end\_header\_id|>} \\[1em]\\[1em]
        $w_c$ &
        \texttt{<|begin\_of\_text|><|start\_header\_id|>system<|end\_header\_id|>}\newline
        Answer the following query considering the provided context. Answer with only one word.\texttt{<|eot\_id|><|start\_header\_id|>user<|end\_header\_id|>}\newline
        Context: Daily Grind Cafe is a popular place to eat lunch out. At a Potluck Party, everyone brings their own lunch to share with friends.\newline
        Instruction: Only consider the context in answering the query.\newline
        Query: Which action can help a person save money?\newline
        A. stop eating at Daily Grind Cafe\newline
        B. attend a Potluck Party\newline
        C. Unknown\texttt{<|eot\_id|><|start\_header\_id|>assistant<|end\_header\_id|>} \\[1em]\\[1em]
        $w_b$ & 
        \texttt{<|begin\_of\_text|><|start\_header\_id|>system<|end\_header\_id|>}\newline
        Answer the following query considering the provided context. Answer with only one word.\texttt{<|eot\_id|><|start\_header\_id|>user<|end\_header\_id|>}\newline
        Context: Daily Grind Cafe is a popular place to eat lunch out. At a Potluck Party, everyone brings their own lunch to share with friends.\newline
        Instruction: Consider the context in answering the query.\newline
        Query: Which action can help a person save money?\newline
        A. stop eating at Daily Grind Cafe\newline
        B. attend a Potluck Party\newline
        C. Unknown\texttt{<|eot\_id|><|start\_header\_id|>assistant<|end\_header\_id|>}\\ \bottomrule
    \end{tabular}
    \caption{Prompt template for intent-driven answer control.}
    \label{tab:app_intent_prompt_template}
\end{table*}

\begin{figure*}[!ht]
\centering

\begin{minipage}{\textwidth}
    \centering
    \begin{tabular}{l|p{13cm}}
        \toprule
        \textbf{Intent} & \textbf{Prompt} \\ \midrule
        NLE &
        \texttt{<|begin\_of\_text|><|start\_header\_id|>system<|end\_header\_id|>}\newline
        Answer the following query considering the provided context. Generate your final answer with only one word. If you are unable to answer the query, generate your final answer as "Unknown". Also, generate an explanation to determine your final answer. Return your output in JSON format: \{\texttt{"explanation": "your explanation here", "answer": "your final response here"}\}. Only include the JSON object in your response.\newline
        \texttt{<|eot\_id|><|start\_header\_id|>user<|end\_header\_id|>}\newline
        Context: Daily Grind Cafe is a popular place to eat lunch out. At a Potluck Party, everyone brings their own lunch to share with friends.\newline
        Query: Which action can help a person save money?\newline
        A. stop eating at Daily Grind Cafe\newline
        B. attend a Potluck Party\newline
        C. Unknown\newline
        \texttt{<|eot\_id|><|start\_header\_id|>assistant<|end\_header\_id|>} \\[1em]

        CoT &
        \texttt{<|begin\_of\_text|><|start\_header\_id|>system<|end\_header\_id|>}\newline
        Answer the following query considering the provided context. Generate your final answer with only one word. If you are unable to answer the query, generate your final answer as "Unknown". Also, generate an explanation to determine your final answer. Return your output in JSON format: \{\texttt{"explanation": "your explanation here", "answer": "your final response here"}\}. Only include the JSON object in your response.\newline
        \texttt{<|eot\_id|><|start\_header\_id|>user<|end\_header\_id|>}\newline
        Context: Daily Grind Cafe is a popular place to eat lunch out. At a Potluck Party, everyone brings their own lunch to share with friends.\newline
        Query: Which action can help a person save money?\newline
        A. stop eating at Daily Grind Cafe\newline
        B. attend a Potluck Party\newline
        C. Unknown\newline
        \texttt{<|eot\_id|><|start\_header\_id|>assistant<|end\_header\_id|>}\newline
        Give your answer by analyzing step by step. \\ \bottomrule
    \end{tabular}
    \caption{Prompt templates for NLE generation and CoT-based prompting.}
    \label{tab:nle_cot_prompt_template}
\end{minipage}

\vspace{1.5em}

\begin{minipage}{0.8\textwidth}
    \centering
    \begin{tabular}{l|c|c|c}
        \toprule
         \textbf{Hyperparameter} & \textbf{Llama} & \textbf{Gemma} & \textbf{Mistral} \\ \midrule
         number of samples & 500 & 200 & 500\\ 
         batch size & 24 & 10 & 10\\
         $\tau_p$ & 0.65 & 0.75 & 0.75\\
         $\tau_c$ & 0.60 & 0.85 & 0.65\\
         margin & 0.3 & 0.3 & 0.3\\
         eps & 0.05 & 0.05 & 0.05\\ \bottomrule
    \end{tabular}
    \caption{Patching hyperparameters for identifying important layers for rank-2 projection subspaces from Llama-3.1-8B-Instruct, Gemma-2-9B-it, and Mistral-7B-Instruct-v0.3.}
    \label{tab:app_patching_hyperparameters}
\end{minipage}

\end{figure*}

\subsubsection{Hyperparameters}\label{app:hyperparameters}
We identify important layers $\mathbb{L}_{b \rightarrow c}$ and $\mathbb{L}_{b \rightarrow p}$ to capture individual PK and CK contributions, respectively, from the final answer $a$ via Patchscope using the hyperparameters described in \cref{tab:app_patching_hyperparameters}.

\subsubsection{Computation Overhead of PatchScope}\label{app:compute_patchscope}
We perform patchscope-based layer selection to learn the rank-2 PK--CK projection subspace on a single NVIDIA B200 GPU (180 GB). For both Llama-3.1-8B and Mistral-v0.3-7B with $32$ layers, it takes less than $30$ minutes for $500$ examples; for Gemma-2-9B with $26$ layers, it takes slightly over $30$ minutes for $200$ examples. This observation verifies negligible computational overhead for patchscope-based layer selection and subsequent rank-2 subspace learning, making it computationally feasible for larger models.

\subsubsection{Evaluation Metric}\label{app:eval_metric}
To empirically verify the insufficiency of the rank-1 projection subspace in disentangling individual knowledge contributions, we utilize two specific metrics:

\noindent \textit{Subspace component:} For each knowledge interaction type, we compute the subspace component $\langle \vec{\mathbf{u}}^T, \vec{\mathbf{h_a}} \rangle$ of the hidden representation of the answer token $\vec{\mathbf{h_a}}$, which is equivalent to the scalar component of the basis vector $\vec{\mathbf{u}}$ for the rank-1 projection subspace $\mathbf{P}$ \citep{minder2025controllable}. If $\mathbf{P}$ adequately captures PK–CK interactions in the dataset, then the distribution of these subspace components should have a mean significantly different from $0$. Such a pattern would indicate minimal contribution in the orthogonal complement $\mathbf{(I - P)}$ and reflect distinct, disentangled interaction behavior across different knowledge types.

\noindent \textit{Explained Variance (Cumulative):} Consider the matrix $H=[\{{\vec{\mathbf{h_{a_j}}}}\}_{j=1}^N], H \in \mathbb{R}^{N \times d}$ as the concatenation of hidden representation $\vec{\mathbf{h_a}} \in \mathbb{R}^d$ of the answer token $a$ over $N$ examples. The $d$ singular values $\sigma_1>>\sigma_2, ...>>\sigma_d$ from the diagonal matrix $\Sigma \in \mathbb{R}^{d \times d}$ after the Singular Value Decomposition (SVD) of $HH^T = A\Sigma B$ indicates the strength of $HH^T$ in $d$ orthonormal directions. For top-$r$ singular values $\sigma_1>>\sigma_2, ...>>\sigma_r$, the cumulative explained variance $EV_r$: $EV_r = \frac{\sum_{j=1}^{r} \sigma_j^{2}}{\sum_{j=1}^{d} \sigma_j^{2}}$

\noindent indicates the sufficiency of rank-$r$ projection subspace in encoding the variance in knowledge interactions present in the dataset \citep{Wall2003,JMLR:v26:24-0699}. If the rank-1 projection subspace $\mathbf{P}$ adequately captures PK--CK interactions in the dataset, then the cumulative explained variance at rank-1 $EV_1$ should approximate $1.0$.

\subsection{Additional Results}\label{app:additional_results}

\begin{theorem}[Non-identifiability under rank-1]
\label{thm:nonidentifiability}
Let the hidden representation $\vec{\mathbf{h_i}}$ for the input $x_i$ at the sequence step $i$ be decomposed as
\[
\vec{\mathbf{h_i}} = c_i \vec{\mathbf{u}}_{CK} + p_i \vec{\mathbf{u}}_{PK} + \boldsymbol{\xi}_i,
\]
where $\vec{\mathbf{u}}_{CK},\vec{\mathbf{u}}_{PK}$ are orthonormal directions corresponding to context and parametric knowledge, $c_i,p_i \in \mathbb{R}$ are their contributions, and $\boldsymbol{\xi}_i$ is noise orthogonal to their span.  
A rank-1 probe with vector $\vec{\mathbf{v}}$ observes
\[
\alpha_i = \vec{\mathbf{v}}^{\top}\vec{\mathbf{h_i}} = c_i \langle \vec{\mathbf{v}},\vec{\mathbf{u}}_{CK}\rangle + p_i \langle \vec{\mathbf{v}},\vec{\mathbf{u}}_{PK}\rangle.
\]
Then $(c_i,p_i)$ are not uniquely identifiable from $\alpha_i$ whenever both coefficients are nonzero.
\end{theorem}

\begin{proof}
Let $a=\langle \vec{\mathbf{v}},\vec{\mathbf{u}}_{CK}\rangle$ and $b=\langle \vec{\mathbf{v}},\vec{\mathbf{u}}_{PK}\rangle$. For any $(c,p)$, choose $(c',p')=(c+\delta,\, p-\tfrac{a}{b}\delta)$ with $\delta \neq 0$.  
Then $ac+bp = ac'+bp'$, so infinitely many $(c',p')$ yield the same observation.  
Thus the mapping $(c,p)\mapsto \alpha$ is non-injective, and the individual contributions cannot be disentangled. 
\renewcommand{\qedsymbol}{}
\end{proof}

\begin{figure}[!ht]
    \centering
    \includegraphics[width=\linewidth]{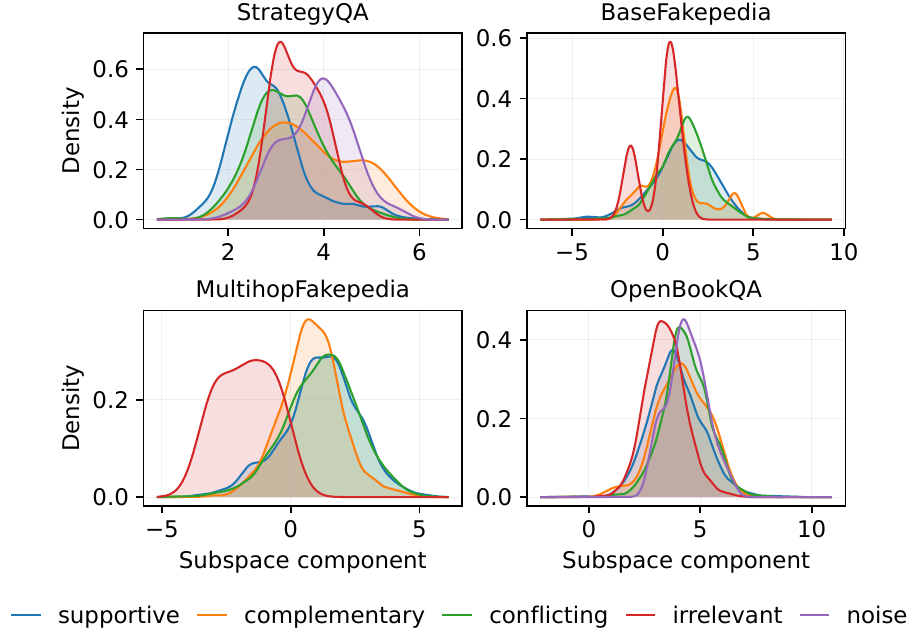}
    \caption{Kernel Density Estimate (KDE) of the PK--CK subspace component 
$\langle \vec{\mathbf{u}}^{T}, \vec{\mathbf{h}}_{i} \rangle$ across different 
knowledge interaction types for four question–answer datasets using the 
gemma-2-9b-it model.}
    \label{fig:rank1_proj_contribution_gemma}
\end{figure}

\begin{figure}[t]
    \centering
    \includegraphics[width=1.0\linewidth]{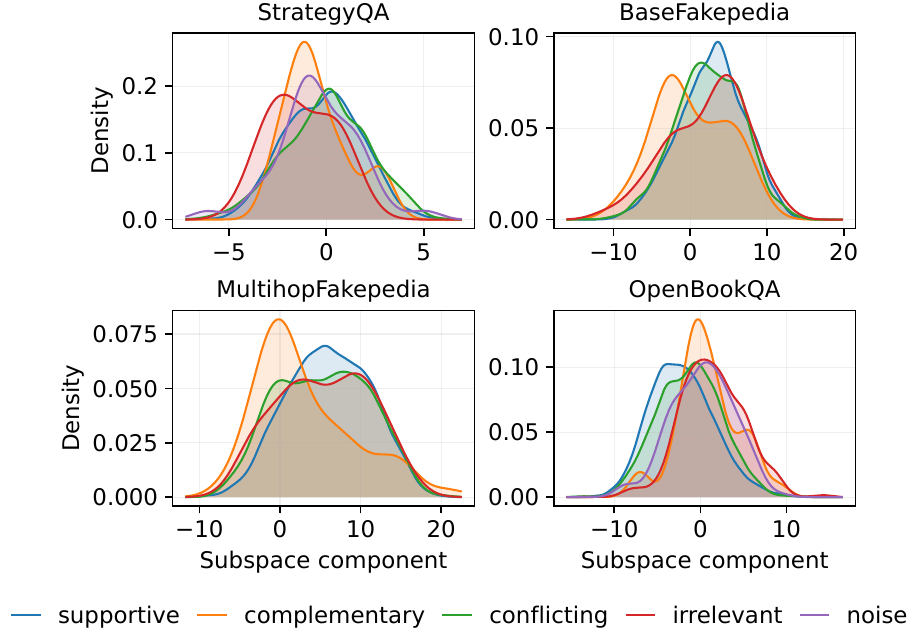}
    \caption{Kernel Density Estimate (KDE) of the PK--CK subspace component $\langle \vec{\mathbf{u}}^{T}, \vec{\mathbf{h}}_{i} \rangle$ across different knowledge interaction types for four QA datasets using Mistral-7B-Instruct-v0.3.}
    \label{fig:rank1_proj_contribution_mistral}
\end{figure}

\begin{figure}[!ht]
  \centering
  \begin{subfigure}{\linewidth}
    \centering
    \includegraphics[width=\linewidth]{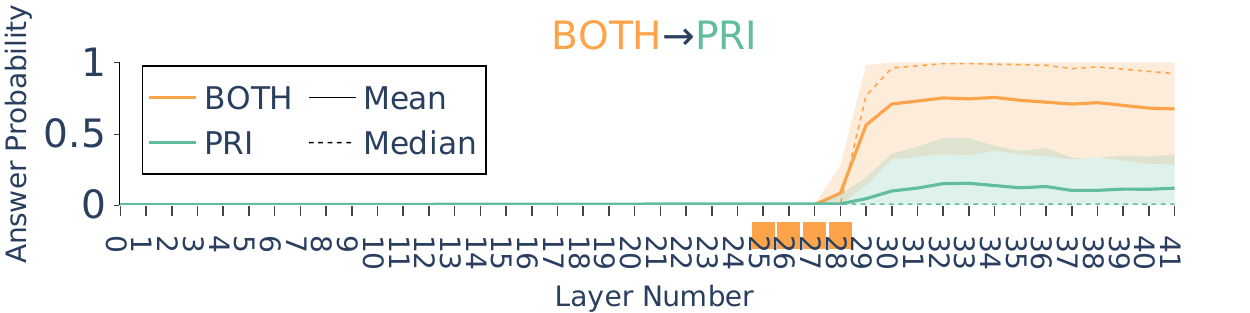}
    \caption{$\mathcal{D}_w^{(b \rightarrow p)}$}
    \label{fig:patchscope_gemma_bp}
  \end{subfigure}\hfill
  \begin{subfigure}{\linewidth}
    \centering
    \includegraphics[width=\linewidth]{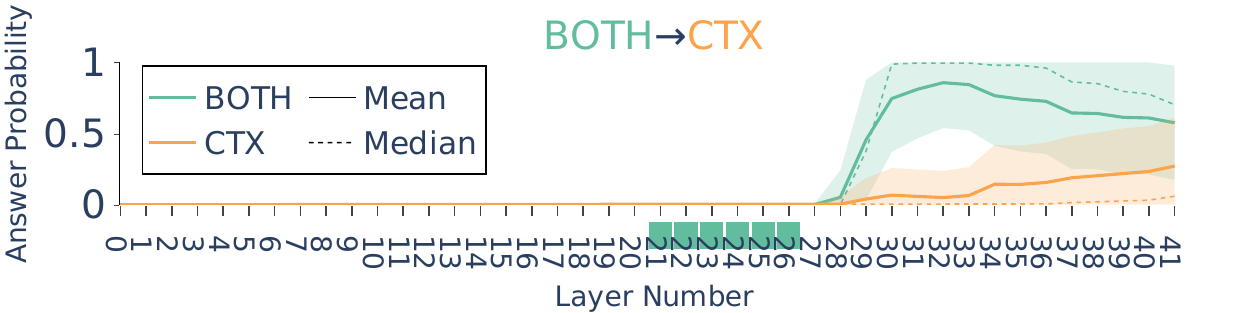}
    \caption{$\mathcal{D}_w^{(b \rightarrow c)}$}
    \label{fig:patchscope_gemma_bc}
  \end{subfigure}

  \caption{Patchscope on OpenBookQA dataset from gemma-2-9b-it.
  (a) Activation patching on $\mathcal{D}_w^{(b \rightarrow p)}$ results in a higher contribution of PK in generating the final answer, as the probability gap between the source and target is higher.
  (b) Activation patching on $\mathcal{D}_w^{(b \rightarrow c)}$ results in a lower contribution of CK in generating the final answer, as the probability gap between the source and target is lower.
  We consider the common layers from both activation patching to learn the rank-2 projection subspace.}
  \label{fig:patching_layer_openbookqa_gemma_2_9b_it}
\end{figure}

\begin{figure}[!ht]
  \centering
  \begin{subfigure}{\linewidth}
    \centering
    \includegraphics[width=\linewidth]{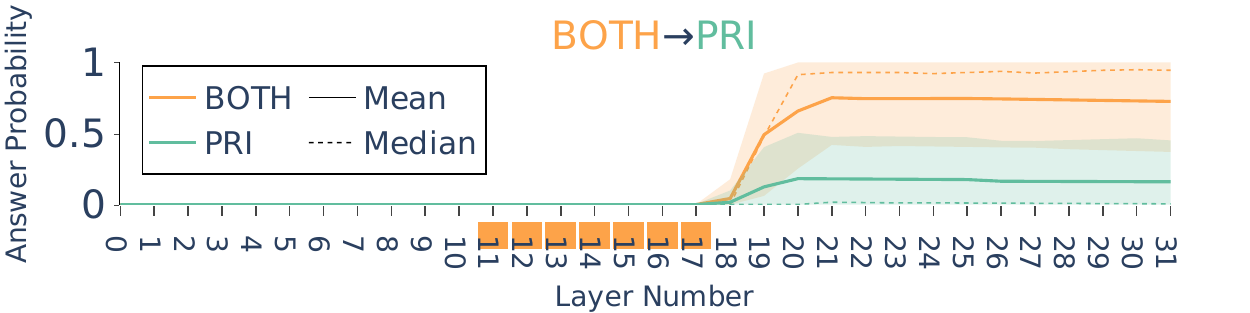}
    \caption{$\mathcal{D}_w^{(b \rightarrow p)}$}
    \label{fig:patchscope_mistral_bp}
  \end{subfigure}\hfill
  \begin{subfigure}{\linewidth}
    \centering
    \includegraphics[width=\linewidth]{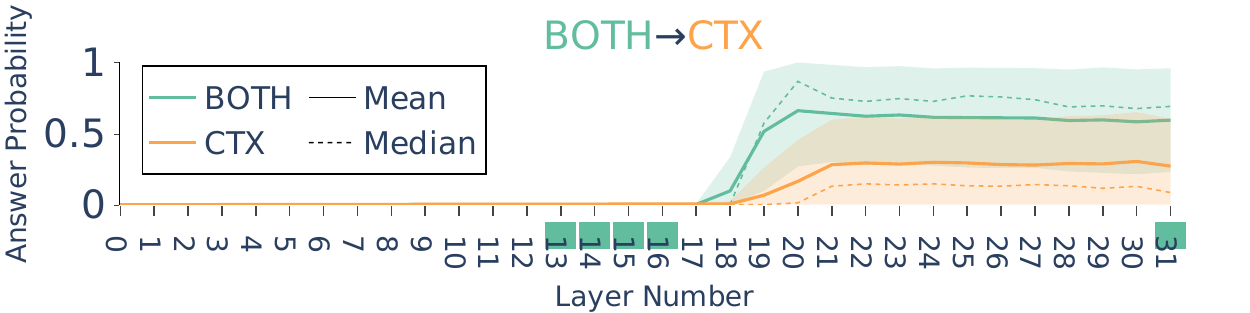}
    \caption{$\mathcal{D}_w^{(b \rightarrow c)}$}
    \label{fig:patchscope_mistral_bc}
  \end{subfigure}

  \caption{Patchscope on OpenBookQA dataset from Mistral-7B-Instruct-v0.3.
  (a) Activation patching on $\mathcal{D}_w^{(b \rightarrow p)}$ results in a higher contribution of PK in generating the final answer, as the probability gap between the source and target is higher.
  (b) Activation patching on $\mathcal{D}_w^{(b \rightarrow c)}$ results in a lower contribution of CK in generating the final answer, as the probability gap between the source and target is lower.
  We consider the common layers from both activation patching to learn the rank-2 projection subspace.}
  \label{fig:patching_layer_openbookqa_mistral_7b_instruct_v0.3}
\end{figure}

\begin{figure*}[t]
  \centering

  \begin{subfigure}{0.48\textwidth}
    \centering
    \includegraphics[width=\linewidth]{./Figures/all_rank2_proj_contrib_pk_ck_dynamics_Meta-Llama-3.1-8B-Instruct_abs.pdf}
    \caption{Overall}
    \label{fig:overall}
  \end{subfigure}\hfill
  \begin{subfigure}{0.48\textwidth}
    \centering
    \includegraphics[width=\linewidth]{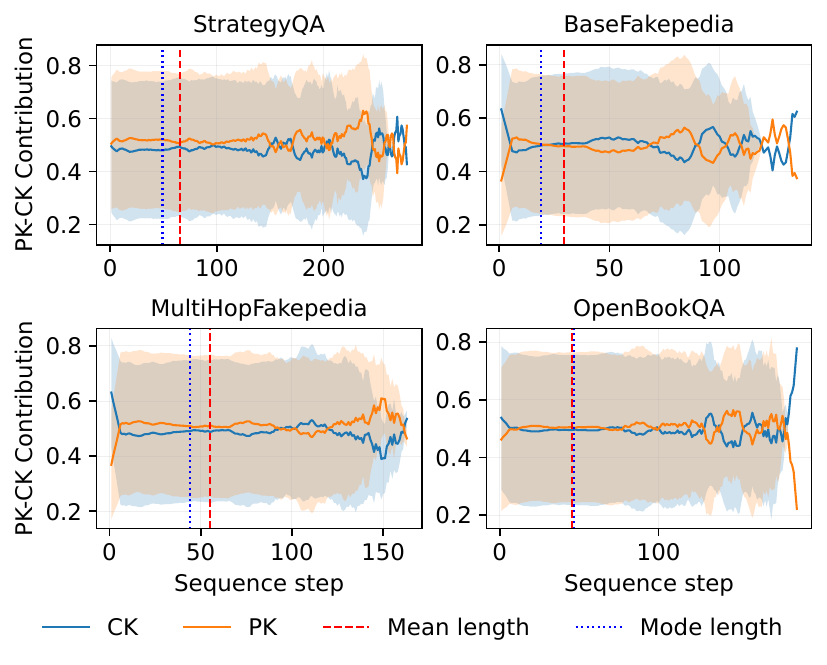}
    \caption{Supportive}
    \label{fig:supportive}
  \end{subfigure}

  \vspace{0.5em}

  \begin{subfigure}{0.48\textwidth}
    \centering
    \includegraphics[width=\linewidth]{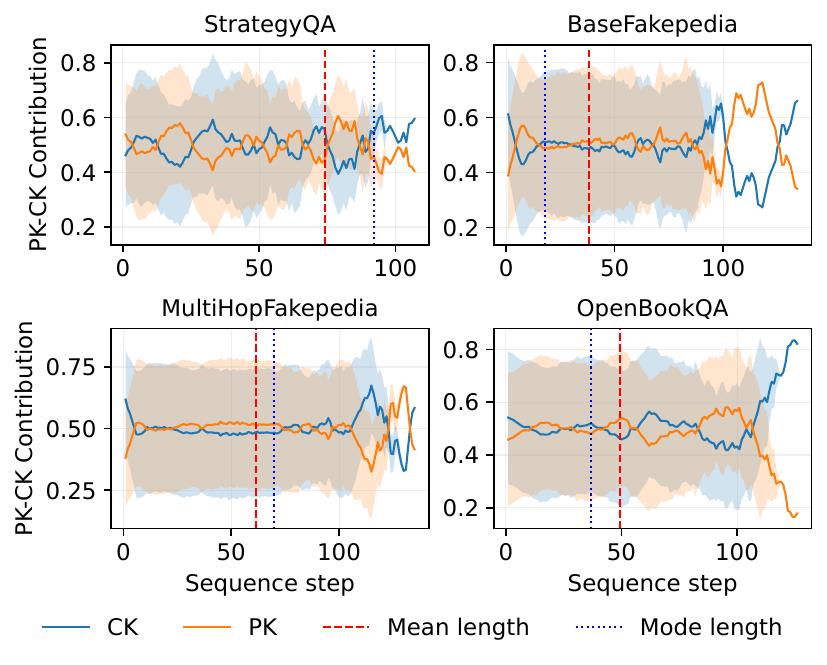}
    \caption{Complementary}
    \label{fig:complementary}
  \end{subfigure}\hfill
  \begin{subfigure}{0.48\textwidth}
    \centering
    \includegraphics[width=\linewidth]{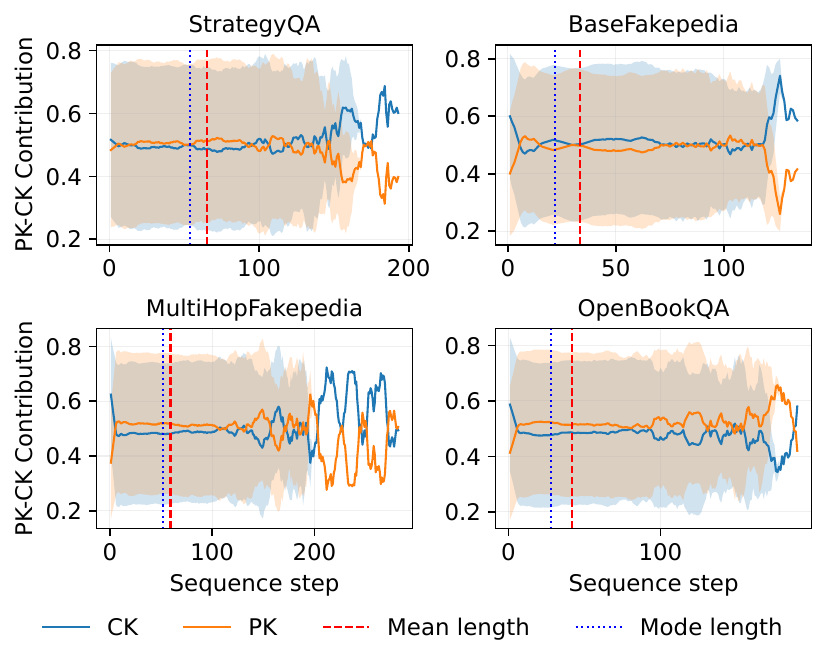}
    \caption{Conflicting}
    \label{fig:conflicting}
  \end{subfigure}

  \caption{PK--CK interaction dynamics over the sequence steps for different interaction scenarios for Llama-3.1-8B-Instruct. The dotted red and blue lines indicate the mean and mode of NLE lengths.}
  \label{fig:pkck_dynamics_llama_detailed}
\end{figure*}

\begin{figure}[!ht]
    \centering
    \includegraphics[width=\linewidth]{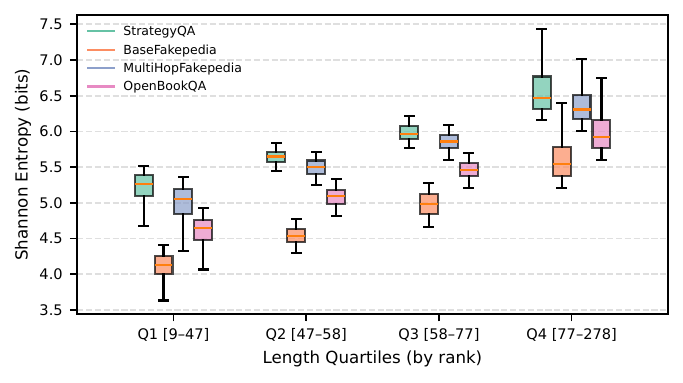}
    \caption{Entropy of NLE generation across all datasets from Llama-3.1-8B-Instruct for different NLE lengths grouped in four quartiles.}
    \label{fig:entropy_vs_nle_quartile}
\end{figure}

\subsection{Robustness Analysis}
\label{app:robustness}

We evaluate the robustness of PatchScope localization and the learned rank-2 PK--CK projection subspace along four axes: layer sensitivity, prompt perturbations, context variation, and geometric stability.

\noindent\textbf{PatchScope layer sensitivity.}
To ensure that PK--CK disentanglement is not tied to a single layer, we evaluate the rank-2 subspace across six PatchScope-identified layers of Llama-3.1-8B-Instruct (L13--L18). Patching accuracy is stable across $6$ layers L13-18,
$[0.82, 0.84, 0.83, 0.85, 0.87, 0.86]$, with a mean of $0.84 \pm 0.018$ and peak performance at L17 ($0.87$). The resulting stability score of $0.98$ indicates that disentanglement consistently emerges across this mid-to-late layer block rather than being a layer-specific artifact.

\noindent\textbf{Prompt perturbation robustness.}
We test invariance to surface-form variations via paraphrasing, synonym substitution, and word-order perturbations. End-to-end prediction agreement remains high (model-dependent), at $0.92$, $0.88$, and $0.85$ for paraphrase, synonym, and word-order changes, respectively. To assess geometric robustness, we independently learn projection subspaces from original and perturbed prompts and measure their cosine similarity, obtaining mean consistency scores of $0.95\pm0.04$, $0.93\pm0.05$, and $0.89\pm0.08$. These results indicate that the learned subspace is largely insensitive to superficial prompt variations and primarily reflects semantic knowledge alignment.

\noindent\textbf{Context variation sanity check.}
If the projection subspace captures genuine knowledge interaction, PK--CK contributions should shift with context availability. Varying context from minimal (no context) to medium (first 50 CK tokens) to maximal (full context) yields:
$\texttt{minimal}: (p_i{=}0.82,c_i{=}0.24)$,
$\texttt{medium}: (p_i{=}0.65,c_i{=}0.58)$,
$\texttt{maximal}: (p_i{=}0.85,c_i{=}0.87)$.
The increase in CK contribution from minimal to medium context reflects greater reliance on evidence as it becomes available, while maximal context strongly activates both knowledge sources, consistent with supportive and complementary interactions rather than pure suppression of PK.

\noindent\textbf{Geometric stability and separation.}
The learned parametric and contextual directions are highly stable across resampling, as shown with examples from the StrategyQA dataset, with average directional similarity exceeding $0.9$ for both axes. Moreover, the two directions remain well separated: the mean cosine overlap between PK and CK axes is $0.00$ across bootstraps. This consistent orthogonality supports the identifiability motivation for a rank-2 formulation.
Overall, these results demonstrate that PK--CK disentanglement is stable across layers, robust to prompt surface perturbations, responsive to context evidence, and geometrically well-conditioned.

\end{document}